\documentclass[pmlr]{jmlr}

\usepackage{longtable}

 \usepackage{booktabs}
 
\usepackage[load-configurations=version-1]{siunitx} 

\usepackage{amsmath}
\usepackage{bbm}

\RequirePackage[nolist,nohyperlinks]{acronym}
\acrodef{ML}{machine learning}
\acrodef{AUROC}{area under the receiver operating characteristic curve}
\acrodef{c-index}{concordance index}
\acrodef{POP}{proportion of patient-pairs}
\acrodef{SGD}{stochastic gradient descent}
\acrodef{DGP}{data generation process}
\acrodef{MIMIC-III}{Multi-parameter Intelligent Monitoring for Intensive Care - III}

\newcommand{\eg}{\textit{e.g.}, }
\newcommand{\ie}{\textit{i.e.}, }

\newcommand{\real}{\mathbbm{R}}

\DeclareMathOperator{\AUROC}{AUROC}

\DeclareMathOperator{\ind}{\mathbbm{1}}
\DeclareMathOperator{\BTC}{\mathcal{C}^{BT}}
\DeclareMathOperator{\RBC}{\mathcal{C}^{R}}

\makeatletter
\def\set@curr@file#1{\def\@curr@file{#1}} 
\makeatother

\jmlrvolume{219}
\jmlryear{2023}
\jmlrworkshop{Machine Learning for Healthcare}

\title[Rank-Based Compatibility for Clinician-Model Teams]{Updating Clinical Risk Stratification Models Using Rank-Based Compatibility: Approaches for Evaluating and Optimizing Clinician-Model Team Performance}

\author{\Name{Erkin \"{O}tle\d{s}}
       \Email{eotles@umich.edu}\\ 
       \addr Medical Scientist Training Program\\
       University of Michigan\\
       Ann Arbor, MI, USA 
       \AND
       \Name{Brian T. Denton}
       \Email{btdenton@umich.edu}\\ 
       \addr Department of Industrial and Operations Engineering\\
       University of Michigan\\
       Ann Arbor, MI, USA
       \AND
       \Name{Jenna Wiens}
       \Email{wiensj@umich.edu}\\ 
       \addr Division of Computer Science and Engineering\\
       University of Michigan\\
       Ann Arbor, MI, USA
       } 

\begin{document}

\maketitle

\begin{abstract}
    As data shift or new data become available, updating clinical \acl{ML} models may be necessary to maintain or improve performance over time. 
    However, updating a model can introduce compatibility issues when the behavior of the updated model does not align with user expectations, resulting in poor user-model team performance.
    Existing compatibility measures depend on model decision thresholds, limiting their applicability in settings where models are used to generate rankings based on estimated risk.
    To address this limitation, we propose a novel rank-based compatibility measure, $\RBC$, and a new loss function that aims to optimize discriminative performance while encouraging good compatibility.
    Applied to a case study in mortality risk stratification leveraging data from MIMIC, our approach yields more compatible models while maintaining discriminative performance compared to existing model selection techniques, with an increase in $\RBC$ of $0.019$ ($95\%$ confidence interval: $0.005$, $0.035$). 
    This work provides new tools to analyze and update risk stratification models used in clinical care.
\end{abstract}

\section{Introduction}

\begin{figure}[!ht]
    \centering
    \includegraphics[width=\textwidth]{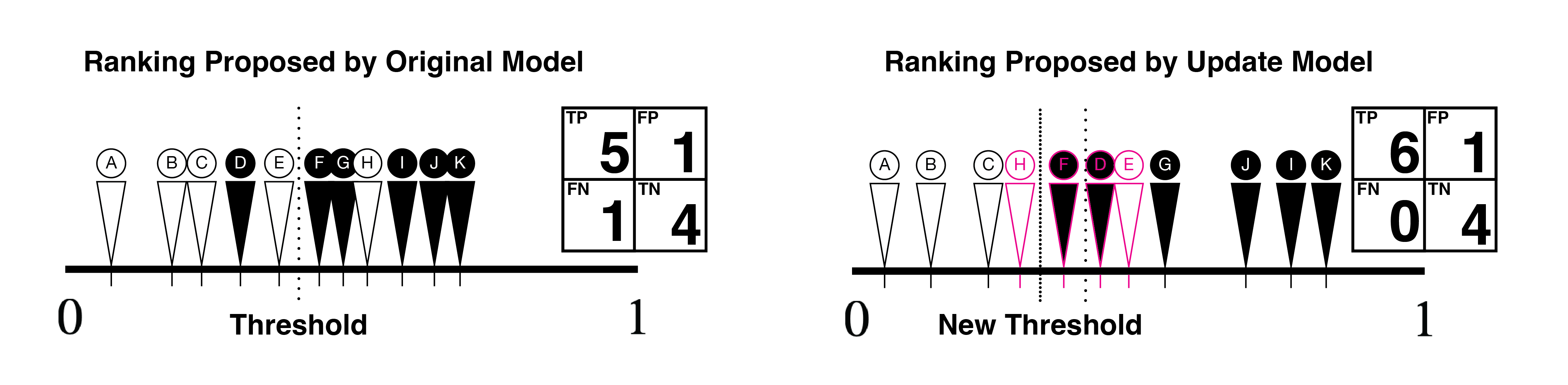}
    \caption[\BTC vs. \RBC Overview]
    {Backwards trust compatibility ($\BTC$, \citet{RN512}) vs. Rank-based compatibility ($\RBC$, proposed).
    A model is used to stratify patients at risk for an outcome (black) from those that are not (white).
    Both the original and updated models have decision thresholds independently set to maximize accuracy on the validation set (shown). On this set, the original model has an accuracy of $\tfrac{9}{11}$ and an \acs{AUROC} of $\tfrac{26}{30}$.
    The updated model switches the order of patients highlighted in magenta, resulting in  higher accuracy $\tfrac{10}{11}$ and \acs{AUROC} $\tfrac{28}{30}$.
    Out of the $9$ patients correctly labeled by the original model the updated model labeled $8$ correctly, this fraction, $\tfrac{8}{9}$, is $\BTC$. This measure depends on the model decision thresholds. Our compatibility measure, $\RBC$, evaluates the ordering of patient-pairs. Of the $26$ patient-pairs correctly ordered by the original model, the updated model correctly ordered $25$ (makes an error on patient-pair E-F), yielding a $\RBC$ of $\tfrac{25}{26}$.
    }
    \label{fig:btc_rbc_overview}
\end{figure}

As \acf{ML} models become increasingly integrated into clinical workflows, understanding the impact of model updates on these workflows and users is crucial.
Models may be retrained and updated as new data become available to maintain or improve performance over time \citep{RN1013, RN1385, RN1386}.
For example, Memorial Sloan Kettering Cancer Center's prostate cancer outcome prediction models are updated annually \citep{RN902}.
While primarily intended to improve model performance, model updating can also affect users' expectations, \ie how users believe a model will perform given specific examples or patients.
When models behave in unexpected ways (\eg make mistakes in situations where they were previously accurate), user-model team performance can suffer \citep{RN512, RN796}.
Thus, selecting updated models based solely on discriminative performance may be insufficient.
Model developers may need to consider the potential disruption to existing workflows and alignment with user expectations in addition to discriminative performance \citep{RN512, RN1336}.
This creates a need for practical tools to estimate how updated models might influence user expectations without directly querying users \citep{RN1338}.
Fundamentally, we would like a way to answer this question: \textit{to what extent does an updated model retain the correct behavior of an original model?}

To this end, \textit{compatibility measures} assess how much an updated model may disrupt a user's mental model compared to the original model and an evaluation dataset.
While researchers have proposed compatibility measures for supervised classification, like the backwards trust compatibility measure, these existing measures depend on a decision threshold \citep{RN512}.
However, selecting a single fixed threshold may not be appropriate in many settings.
In the context of patient risk stratification tools, decision thresholds can depend on system constraints or user preferences \citep{RN1335,RN1326}.
Similar to how the receiver operating characteristic curve evaluates discriminative performance across all decision thresholds, there is a need for compatibility measures that are independent of a threshold.

\begin{figure}[!ht]
    \centering
    \includegraphics[width=\textwidth]{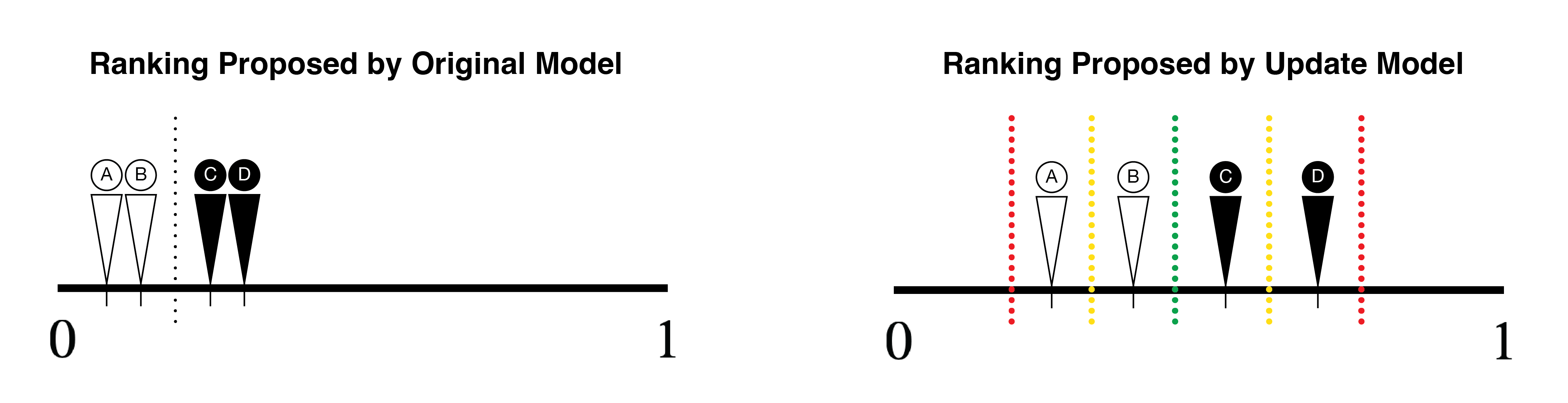}
    \caption[Threshold Dependence of $\BTC$]
    {$\BTC$ is sensitive to the choice of both model decision thresholds.
    Both models have perfect rank-based discrimination (\ie $\AUROC=1$).
    Depending on the updated decision threshold, $\BTC$ may be $\tfrac{1}{2}$ (red), $\tfrac{3}{4}$ (yellow), or $1$ (green). Regardless of the model decision threshold, $\RBC$ is $1$ for this example.
    }
    \label{fig:btc_various_thresholds}
\end{figure}

Given this gap, we propose a novel rank-based compatibility measure that estimates the probability that an updated model will correctly rank a pair of discordantly labeled patients (a \textit{patient-pair}), given that the original model was correct.
This new measure offers a broader evaluation framework for model updates used in risk stratification and ranking, and applies in settings where model outputs are used for clinical resource allocation decisions. By considering the concordance between model rankings, we can proactively detect potentially harmful updates and avoid negative impacts on user-model team performance.
\textbf{Figure \ref{fig:btc_rbc_overview}} provides an overview of our proposed approach, illustrating its relationship to existing performance and compatibility measures. \textbf{Figure \ref{fig:btc_various_thresholds}} illustrates the limitations of backwards trust compatibility compared to rank-based compatibility.
In this work, we also demonstrate how our new measure relates to model discriminative performance and develop a loss function that can be used to directly optimize for compatibility.

\subsection*{Generalizable Insights about Machine Learning in the Context of Healthcare}

Healthcare has witnessed an explosion of \ac{ML} models in recent years, and it is a domain in which the task of ranking patients based on risk arises frequently.
At the same time, models must be updated to retain clinical utility.
For example, the Epic sepsis model, a patient deterioration model used by tens of thousands of clinicians in the United States, was recently updated in light of reports of poor performance \citep{RN1390, RN982, RN1387}.
We focus on a similar case study in which we stratify patients according to their risk of in-hospital mortality.
While it may seem that discriminative performance must suffer to maintain compatibility, we show that developers can generate compatible updated models without negatively affecting discriminative performance by using our proposed loss function during training. 
Compared to updating approaches that ignore compatibility, or use existing compatibility measure, this work facilitates model updates that are more consistent with clinicians' expectations and thus may be more readily accepted and adopted in practice.\\

\noindent
Our main contributions are as follows:

\begin{itemize}

    \item To the best of our knowledge, we introduce the first rank-based compatibility measure based on the concordance of risk estimate pairs.
    
    \item We characterize the extent to which the new compatibility measure may vary over potential model updates.
    
    \item We introduce a loss function that incorporates ranking \textit{incompatibility loss}, which can be used to train model updates with improved rank-based compatibility characteristics.
    
    \item Using MIMIC-III, we present empirical results that demonstrate how the proposed loss function leads to improved rank-based compatibility without a significant decrease in AUROC compared to standard model selection approaches.

\end{itemize}

\section{Problem Setup \& Background}

In the context of learning risk stratification models, a patient $i$ is represented by the tuple $(\mathbf{x}_i,y_i)$, where $\mathbf{x}_{i} \in \real^d $ represents the feature vector and $y_i \in \{0,1\}$ represents the binary label (\eg outcome).
Risk stratification model, $f(\cdot)$, outputs risk estimates, $\hat{p}_i \in [0, 1]$ that estimate $\Pr(y_i=1 | \mathbf{x}_{i})$.
These risk estimates can be converted to predicted labels, $\hat{y}_i = \ind( \hat{p}_i > \tau )$, where $\tau$ is some decision threshold.

We seek to assess the impact on user expectations when updating an original model, $f^o(\cdot)$, to an updated model, $f^u(\cdot)$.
Note that the original and updated models are specific instantiations of the risk stratification models introduced above.
They produce risk estimates denoted as $\hat{p}_i^o$ and $\hat{p}_i^u$, respectively.
We refer to the combination of an original and updated model as a \textit{model-pair}.
Decision thresholds for the original and the updated models are $\tau^o$ and $\tau^u$, respectively.

The original and candidate updated risk stratification models are evaluated on a held-out set of patients, denoted as $I$.
This set can be partitioned into two mutually exclusive subsets based on patient labels: $0$-labeled patients, $I^0$, and $1$-labeled patients, $I^1$.
The size of these subsets of patients are denoted as $n^0$ and $n^1$, respectively, and their sum, $n$, is the cardinality of $I$.
We formalize the notion of a \textit{patient-pair}, a pair of patients $i$ and $j$ that do not share the same label (\ie $i \in I^0$ and $j \in I^1$).
The total number of patient-pairs, $m$, is the product $n^0 n^1$.
We denote the number of patient-pairs correctly ranked by the original and updated models as $m^{o+}$ and $m^{u+}$, respectively.
Both $m^{o+}$ and $m^{u+}$ are integers taking on values between $0$ and $m$ inclusively.
Given an original model, we aim to select an updated model that achieves good discriminative performance and compatibility.

\subsection{Discriminative Performance}

Discriminative performance measures a model's ability to separate patients with different labels \citep{RN1168}.
The \ac{AUROC} is widely used to evaluate the discriminative performance of risk stratification models since it evaluates performance across all decision thresholds $\tau$.
The \ac{AUROC} corresponds to the probability of correctly ranking two patients with differing labels based on the risk estimates produced by the model.
It may be estimated by counting the number of patient-pairs ranked correctly by a model, $m^{o+}$, and then normalizing by the total number of patient-pairs $m$ \citep{RN711}:

\begin{equation}
    \AUROC(f^o) = 
    \frac{
        \sum \limits_{i \in I^0}
            \sum \limits_{j \in I^1}
                \ind(\hat{p}_i^o < \hat{p}_j^o)
        }{
        m
    }
    =
    \frac{
        m^{o+}
        }{
        m
    }
\end{equation}

\noindent
The \ac{AUROC} ranges between $0$ and $1$; a value of $0.5$ corresponds to an ordering that is no better than random.
The \ac{AUROC} is the binary case of the \ac{c-index}, and both are related to the Wilcoxon-Mann-Whitney U statistic \citep{RN1167, RN1306, RN1168}.

\subsection{Backwards Trust Compatibility}
\label{sec:backwards_trust_compatibility}
Currently, \textit{backwards trust compatibility} ($\BTC$) is the primary compatibility measure described in the literature \citep{RN512, RN514}.
$\BTC$ measures the agreement between the true label and the predicted labels produced by the original and updated models by counting the number of patients both labeled correctly and normalizing by the number of patients the original model labeled correctly:

\begin{equation}
    \BTC(f^o, f^u) =
    \frac{
        \sum \limits_{i \in I}
            \ind(y_i = \hat{y}_i^o)
            \cdot
            \ind(y_i = \hat{y}_i^u)
        }{
        \sum \limits_{i \in I}
            \ind(y_i = \hat{y}_i^o)
    }
\end{equation}

\noindent
$\BTC$ depends on an evaluation set of patients, $I$, and values range between $0$ and $1$.
$\BTC=0$ when the updated model mislabels all the patients labeled correctly by the original model, and $\BTC=1$ when the updated model correctly labels all the patients the original model got correct. 
$\BTC$ is not symmetric, as $\BTC(f^o, f^u)$ does not necessarily equal $\BTC(f^u, f^o)$.
$\BTC$ is expected to decrease in settings with dataset shifts as the feature-label relationships captured by the model-pairs differ \citep{RN507}.

In the context of patient risk stratification models, calculating $\BTC$ requires first thresholding risk scores to produce binary predictions.
However, many settings in healthcare do not use a decision threshold \citep{RN1335}.
For example, patients in the emergency department may be stratified by continuous risk estimates, and surgeons may use different risk thresholds to recommend surgery. 
In use cases where there are multiple thresholds, $\BTC$ may be computed multiple times; however, this is problematic for several reasons.
First, the evaluation grows proportionally with the number of thresholds being considered.
Second, there is limited utility in doing so for cases with a class imbalance (see \textbf{Appendix Section \ref{sec:btc_across_thresholds}})
Third, $\BTC$ is sensitive to the selection of thresholds, and poorly chosen thresholds could lead to a model with good discrimination being evaluated poorly, as shown in \textbf{Figure \ref{fig:btc_various_thresholds}}.
These suggest a need for a compatibility measure that applies directly to continuous risk estimates without thresholding.

\section{Methods}

We present our proposed rank-based compatibility measure, $\RBC$, which measures compatibility independent of a decision threshold by examining the ranking concordance of patient-pairs.
While related to the \ac{AUROC}, we hypothesize that optimizing discriminative model performance by minimizing binary cross-entropy loss when training models may not necessarily lead to high $\RBC$. Thus, we propose a new loss function based on a differentiable approximation of $\RBC$ that can be used when training updated models.

\subsection{Rank-Based Compatibility}

The rank-based compatibility, presented in \textbf{Equation \ref{eq:rbc}}, compares the ranking produced by the updated model against the ranking produced by the original model.

\begin{equation}
    \label{eq:rbc}
        \RBC(f^o, f^u) :=
        \frac{
            \sum \limits_{i \in I^0}
                \sum \limits_{j \in I^1}
                    \ind(\hat{p}_i^o < \hat{p}_j^o)
                    \cdot
                    \ind(\hat{p}_i^u < \hat{p}_j^u)
            }{
            \sum \limits_{i \in I^0}
                \sum \limits_{j \in I^1}
                    \ind(\hat{p}_i^o < \hat{p}_j^o)
        } 
\end{equation}

\begin{table*}[!t]
  \centering 
  \caption
  {Relationship between original and updated model \ac{AUROC}, proportion of patient-pairs and count variables.}
  \begin{tabular}{l|ll|l}
  \toprule
                & \textbf{Original Model} & \textbf{Original Model} &  \\
                & \textbf{Ranks Correctly} & \textbf{Ranks Incorrectly} & \\
    \hline
    \textbf{\begin{tabular}[c]{@{}l@{}}Updated Model\\ Ranks Correctly\end{tabular}}  
        & $\phi^{++} = \frac{m^{++}}{m}$ & $\phi^{-+} = \frac{m^{-+}}{m}$ & $\AUROC(f^u) = \frac{m^{u+}}{m}$ \\ 
    & & &  \\ 
    \textbf{\begin{tabular}[c]{@{}l@{}}Updated Model\\ Ranks Incorrectly\end{tabular}}  
        & $\phi^{+-} = \frac{m^{+-}}{m}$ & $\phi^{--} = \frac{m^{--}}{m}$ & $1-\AUROC(f^u)$ \\ 
    \hline
                & $\AUROC(f^o) = \frac{m^{o+}}{m}$ & $1-\AUROC(f^o)$ &  \\
    \bottomrule
  \end{tabular}
  \label{tab:pop_table} 
\end{table*}

\noindent
Given a set of evaluation patients, $I$, $\RBC$ corresponds to the number of patient-pairs that both models rank correctly normalized by the number of patient-pairs that the original model ranked correctly. 
In contrast with $\BTC$, which operates by counting patients, $\RBC$ operates on patient-pairs produced by the mutually disjoint subsets $I^0$ and $I^1$.
$\RBC$ measures the concordance of ranking patient-pairs and ranges from $0$ to $1$.
In contrast, $\BTC$ measures concordance with respect to binary patient predictions.

Although this work focuses on risk stratification models that operate over patients with binary outcomes, $\RBC$ is not limited to this setting; we present a general form of $\RBC$ in \textbf{Appendix Equation \ref{eq:gf_rbc}}.

\paragraph{Relationship to \ac{AUROC}.} Both $\RBC$ and \ac{AUROC} involve counting correct patient-pair rankings.
We introduce several ancillary rank-based compatibility variables to clarify how $\RBC$ relates to \ac{AUROC}.
Four \ac{POP} variables measure how two models rank (correctly vs. incorrectly) patient-pairs.

The \ac{POP} variables, $\phi^{ab}$, follow a convention where $a$ represents how the original model ranks patient-pairs correctly ($+$) vs. incorrectly ($-$), and $b$ represents the same information for the updated model.
For example, the \ac{POP} variable for patient-pairs \textit{correctly} ordered by both models is denoted by $\phi^{++}$, and the proportion of patient-pairs \textit{incorrectly} ordered by both models is $\phi^{--}$.
The four \ac{POP} variables sum to $1$.
From the \ac{POP} variables, one can calculate the \ac{AUROC} of each model (\eg $\AUROC(f^o) = \phi^{++} + \phi^{+-}$).
Each \ac{POP} variable is proportional to a patient-pair count variable: $m^{++}, m^{+-}, m^{-+}$, and $m^{--}$, which follow the same $\cdot^{ab}$ notation.
The relationships among the \ac{POP} variables, the count variables, and discriminative performances can be expressed in a tabular manner, as depicted in \textbf{Table \ref{tab:pop_table}}. From these relationships, $\RBC = \frac{m^{++}}{m^{o+}} = \frac{\phi^{++} }{\AUROC(f^o)}$.

\paragraph{Rank-Based Compatibility Lower Bound.}
Given $\AUROC(f^o)$ and $\AUROC(f^u)$, we can bound all \ac{POP} variables (see Appendix Section \ref{sec:app_pop_variable_bounds}).
Here, we assume that $0.5 < \AUROC(f^o) \leq \AUROC(f^u) \leq 1$, yielding the following lower bound for the rank-based compatibility:

\begin{equation*}
\label{eq:rbc_bounds}
    \tfrac{
        \AUROC(f^o) + \AUROC(f^u)-1
    }{
        \AUROC(f^o)
    } 
    \le \RBC(f^o, f^u) 
\end{equation*}

\noindent 
This bound can be used to contextualize the $\RBC$ of an update, as the range of $\RBC$ changes depending on the model \acp{AUROC} being considered. The lower bound of $\RBC$ increases with respect to the \ac{AUROC} of the updated model (shown graphically in \textbf{Appendix Section \ref{sec:app_lower_bound_of_rbc}}).
We note that the upper bound is always $1$ for the model updating region we are interested in.


\subsection{Optimizing for Rank-Based Compatibility}
\label{sec:incompatibility_loss}

\newcommand{\lbce}{\mathcal{L}^{BCE}}
\newcommand{\lrbc}{\mathcal{L}^{R}}
\newcommand{\approxlrbc}{\widetilde{\lrbc}}

While standard model training and selection procedures that typically focus on discriminative performance will result in a larger lower bound for $\RBC$, one may choose to optimize directly for $\RBC$.
However, as defined, $\RBC$ is non-differentiable due to the \textit{ranking indicator function}, $\ind(\hat{p}_i < \hat{p}_j)$.
To facilitate the use of rank-based incompatibility loss in gradient-based optimization, we introduce a differentiable approximation of rank-based compatibility:

\begin{equation*}
    \widetilde{\RBC}(f^o, f^u) =
    \frac{
        \sum \limits_{i \in I^0}
            \sum \limits_{j \in I^1}
                \sigma(\hat{p}_j^o - \hat{p}_i^o)
                \cdot
                \sigma(\hat{p}_j^u - \hat{p}_i^u)
        }{
        \sum \limits_{i \in I^0}
            \sum \limits_{j \in I^1}
                \sigma(\hat{p}_j^o - \hat{p}_i^o)
    }
\end{equation*}

\noindent
This approximation replaces the ranking indicator function used to evaluate patient pairs with a \textit{ranking sigmoid function}:
\begin{equation*}
    \sigma( \hat{d}_{ji} ) = \frac{1}{1+\exp(-s \cdot \hat{d}_{ji} )}
\end{equation*}

\noindent
Where $\hat{d}_{ji}$ is the difference in risk estimates produced for a patient pair (\ie $\hat{d}_{ji}= \hat{p}_j - \hat{p}_i$ and ranges between $-1$ and $1$).
A correct ranking corresponds to $\hat{d}_{ji} > 0$ and an incorrect ranking corresponds to $\hat{d}_{ji} <0$.
The sigmoid function maps this to a value between $0$ and $1$, closer to the behavior of the ranking indicator function \citep{RN1304}.
A hyperparameter, $s$, controls the spread of this mapping.
Note that using a sigmoid to overcome discontinuity in the loss function is similar to work introduced to optimize for the \ac{AUROC} directly \citep{RN1240}.

Risk stratification models are often trained by minimizing the binary cross-entropy loss $\lbce$. This attempts to optimize the discriminative performance of the model by reducing the probability estimates for $0$-labeled patients and increasing them for $1$-labeled patients, and indirectly optimizes the correct ranking of patient-pairs, the \ac{AUROC} \citep{RN1225}.
However, $\lbce$ only examines the relationship between a patient's label and the risk estimates produced by a model.
To incorporate rank-based compatibility, we augment model update training to incentivize rank-based compatibility, using a weighted combination of binary cross-entropy and $\approxlrbc=1-\widetilde{\RBC}$:

\begin{equation}
    \label{eq:objective_fx}
    \alpha\lbce + (1-\alpha)\approxlrbc \text{ where } \alpha \in [0,1]    
\end{equation}

\noindent
Hyperparameter $\alpha$ controls the trade-off between discriminative performance and compatibility.
During training, the predictions produced by the original model are incorporated into the loss function.

\section{Experiments \& Results}

We focus on understanding and engineering model updates in terms of $\RBC$ using a real-world benchmark dataset.
While $\RBC$ could be used as a validation metric when selecting among candidate models during an update procedure, we hypothesize that by including $\widetilde{\RBC}$ in the loss function, we can achieve better compatibility without paying a penalty in terms of AUROC.
To test this hypothesis, we generated and analyzed model updates on the MIMIC-III mortality prediction dataset.

\paragraph{Questions.}
Our experiments seek to answer two related questions:
\begin{enumerate}
\item \textit{What is the empirical distribution of $\RBC$ achieved using standard model updates when using real data?
}
(%
    \textbf{Section \ref{sec:chpt4_ear_rbc_central_tendency}},
    \textbf{Figure \ref{fig:mortality_update_compatibility}}%
)

\item \textit{Compared to standard model update generation and selection approaches, can we use the rank-based incompatibility loss, $\approxlrbc$, to generate updates with better $\RBC$?
Can this be accomplished without a loss of \ac{AUROC}?
}
(%
    \textbf{Section \ref{sec:chpt4_ear_engineered_loss_vs_standard_selection}},
    \textbf{Figures \ref{fig:selection_vs_engineering_deltas_alpha_beta}}, 
    \textbf{\ref{fig:selection_vs_engineering_deltas_alpha_fixed_0.6}},
    \textbf{\ref{fig:selection_vs_engineering_deltas_beta_fixed_0.6}}, 
    \textbf{\ref{fig:selection_vs_engineering_improvement_statistically_significant}}, and 
    \textbf{\ref{fig:selection_vs_engineering_improvement_breakdown}}%
)
\end{enumerate}

\subsection{Data \& Model Updating Setup}

\paragraph{Dataset \& Task.}
We use \citet{RN512}'s work as foundation for our experimental setup. 
Their experimental work analyzing $\BTC$ in the setting of updating an in-hospital mortality prediction model served as a template for our main analyses.
In order to maintain consistency and enable comparisons between $\BTC$ and $\RBC$ we modeled our predictive task, dataset splits, and model architectures considered off of their initial experiments.

Like \citet{RN512}, we employed the MIMIC-III dataset \citep{RN723}, with the goal of predicting in-hospital mortality based on the first 48 hours of a patient's ICU stay, with the population and task defined by \citet{RN724}.
The data were transformed using FIDDLE \citep{RN807}.
For details regarding the data inclusion and transformation, please see the procedures detailed by \citet{RN807}.
Since our goal wasn't to learn the best possible mortality prediction tool, but to investigate the applicability of $\RBC$, we reduced the number of features from $350,832$ to $35,000$, 
for computational efficiency.
This was done by random sampling.

\begin{figure}[!ht]
    \centering
    \includegraphics[width=\columnwidth]{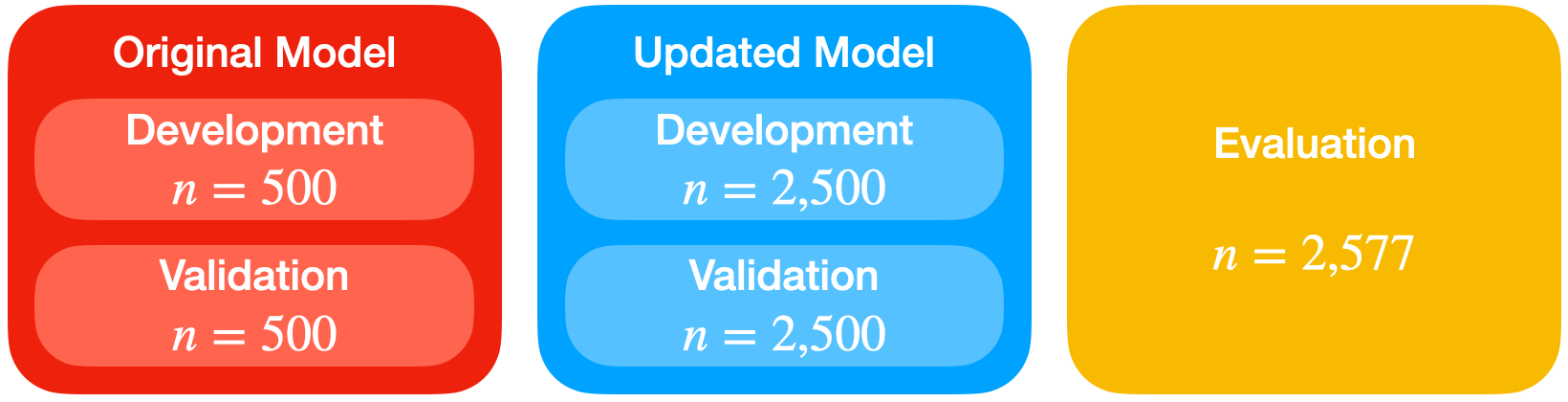}
    \caption[Dataset Splits]
    {The MIMIC-III mortality data was partitioned into three datasets.
    Two of these datasets were allocated for model development and validation, and one was held-out for evaluation.
    Model-pairs were evaluated on the evaluation dataset.
    }
    \label{fig:dataset_splits}
\end{figure}

We randomly split the MIMIC-III data into three disjoint datasets.
Two of these datasets were allocated for model development and validation.
The third dataset was reserved for held-out evaluation.
$8,577$ patients in the MIMIC-III dataset meet the in-hospital mortality inclusion criteria defined by \citet{RN724}.
The datasets were split similarly to \citet{RN512}, with $1,000$  allocated to the original model dataset, $5,000$ were assigned to the updated model dataset, and $2,577$ held-out for the evaluation dataset.
The two model datasets were used to develop and validate the original and updated models.
The model datasets were each split equally (50/50\%) into development and validation datasets.
The dataset partitions and their sizes are depicted in \textbf{Figure \ref{fig:dataset_splits}}.

\paragraph{Original model training \& selection.}
Original models were trained using regularized logistic regression. L2 regularization strength $\{0.1, 0.01, 0.001\}$ was selected to maximize validation \ac{AUROC}.

\paragraph{Updated model training \& selection.}
Two different types of updated models were created to assess standard updating approaches against our proposed loss function. Standard updates, ``BCE models'', were trained to minimize $\lbce$ subject to regularization. The same regularization weights used for the original models were available for the updated models. 


Using the same original model and data, we generated additional updated models, ``RBC models'' based on a loss function that incorporates $\lrbc$ (\textbf{Equation \ref{eq:objective_fx}}), sweeping $\alpha$ in the set $\{ 0, 0.1, 0.2, ..., 0.9, 1\}$. 

Updated models from the ``BCE" and ``RBC models" were selected based on maximizing the following validation function:
\begin{equation}
    \label{eq:selection_fx}
    \beta \AUROC(f^u) + (1-\beta) \RBC(f^o, f^u) \text{ where } \beta \in [0,1]
\end{equation}

\paragraph{Evaluation.}
The selected updated models were evaluated in terms of $\RBC$ and \ac{AUROC} on the held-out evaluation dataset.
The process of splitting the data, training model-pairs, and evaluation was replicated $40$ times.

\subsection{Rank-Based Compatibility Distribution}
\label{sec:chpt4_ear_rbc_central_tendency}

We first investigate: \textit{
What is the empirical distribution of $\RBC$ achieved using standard model updates (\ie minimizing the binary cross-entropy loss) when using real data?
}
Using the experimental setup described above, we created $150$ standard updated models for each original model, minimizing $\lbce$.
To introduce variation, these $150$ candidate ``BCE models'' were created by combining dataset resampling, shuffling, and regularization weights.
The updated model development dataset was either resampled with replacement ($45$ of the times) or shuffled ($5$ of the times, which yields difference in models due to our use of \acl{SGD}) and then models were trained using binary cross-entropy loss with one of three L2 regularization weights ($(45+5) \cdot 3 = 150$).

\begin{figure}[!ht]
    \centering
    \includegraphics[width=\columnwidth]{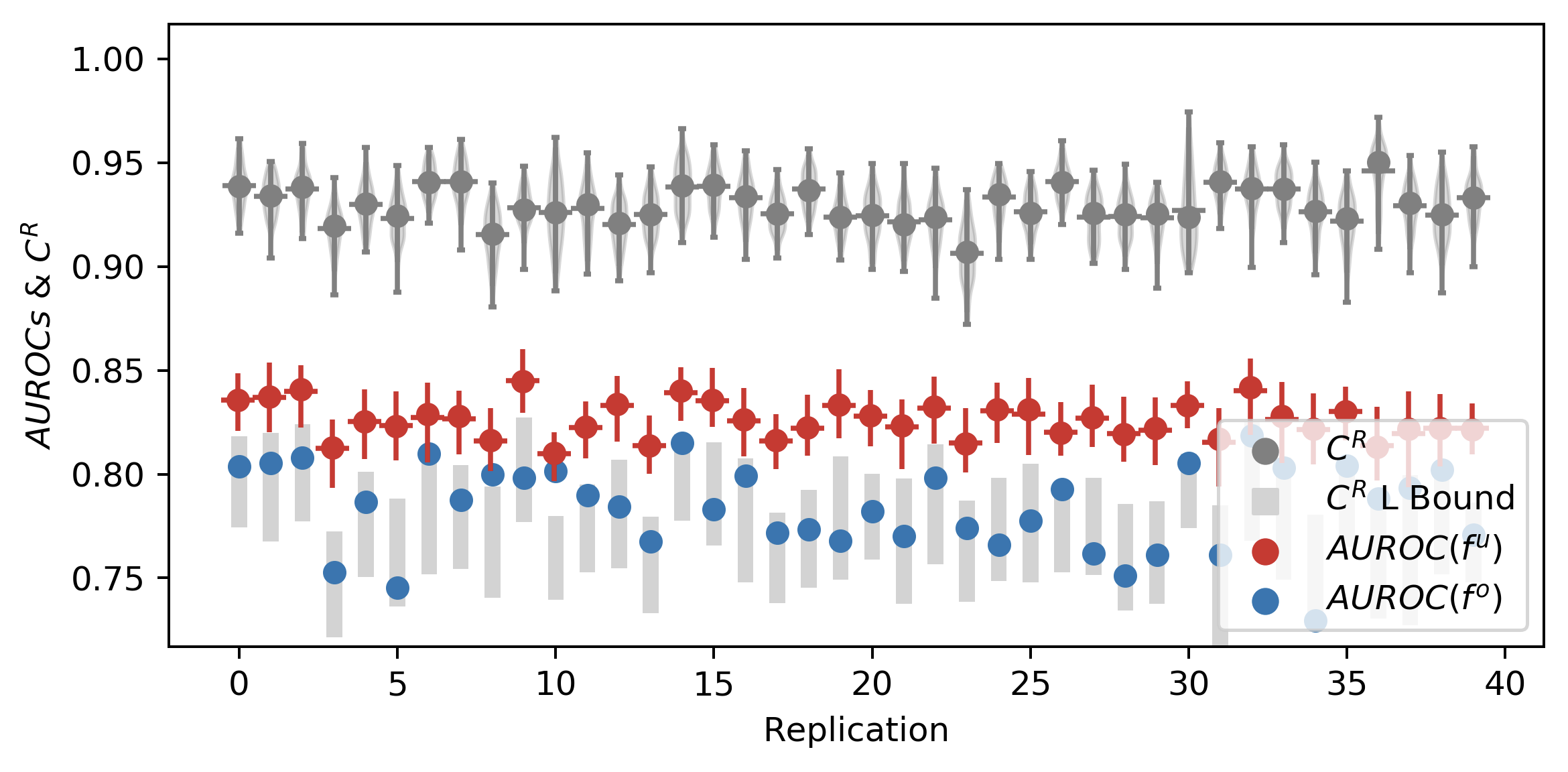}
    \caption[$\RBC$ Distribution For Model Updates on the MIMIC-III Mortality Task]
    {$\RBC$ Distribution For Model Updates on the MIMIC-III Mortality Task.
    An original model was selected for each replication and $150$ ``BCE models'' were generated as candidate updates.
    We plot the \ac{AUROC} of the original model (blue dots) and updated ``BCE models'' (red, $95\%$ confidence intervals).
    We also show the expected lower bounds for $\RBC$ (light gray).
    Finally, the ``BCE models'' $\RBC$s distribution are plotted as violin plots (gray).
    }
    \label{fig:mortality_update_compatibility}
\end{figure}

We calculated the \ac{AUROC} of the original model and the resultant $\RBC$ and \ac{AUROC} across the candidate update models (\textbf{Figure \ref{fig:mortality_update_compatibility}}).
Across the $150$ ``BCE models,'' empirical $95\%$ confidence intervals were calculated for $\AUROC(f^u)$, and violin plots were generated for $\RBC$.

The observed $\RBC$ values for the set of candidate updates vary across a portion of the feasible range (between the lower%
\footnote{Note, that the lower bound is presented as a range.
This is because each candidate update model has a separate lower bound depending on its \ac{AUROC}.}
and upper bounds).
We note that the observed distributions of $\RBC$ shifts in relation to the \acp{AUROC} of the models.
To control for this shift, we can examine the \ac{POP} variable $\phi^{++}$.
We see that the distribution of $\phi^{++}$ for this experiment tends to a central value; this is shown and discussed in \textbf{Appendix Section \ref{sec:phi_plus_plus_central_tendency}}.
These results show the behavior of $\RBC$ for one data-generating process, where we see some variation in $\RBC$ values that provide limited options for model developers to select among.
Additionally, we see that larger $\RBC$ values are possible but not observed through standard update generation procedures (this is the space above the observed $\RBC$ violin plots in \textbf{Figure \ref{fig:mortality_update_compatibility}}).
These findings are underscored in an analytical sketch discussed in \textbf{Appendix Section \ref{sec:app_central_tendency_of_rbc}}.
\textit{All together, these results mean that model developers may be constrained if they wish to develop updated models that optimize for $\RBC$ using standard update generation procedures.}

\subsection{Weighted Loss vs. Standard Updated Model Selection}
\label{sec:chpt4_ear_engineered_loss_vs_standard_selection}

We now investigate our second question: 
\textit{Compared to optimizing for $\lbce$ alone, does incorporating the rank-based incompatibility loss, $\lrbc$, generate updates with better $\RBC$?}

For each replication, we generated $150$ ``BCE models'' using the generation procedure described above.
For each value of $\alpha \in \{0,0.1,0.2, ... ,0.9, 1\}$, we also generated $3$ ``RBC models''.
This was done by sweeping the regularization strengths used above.
Aside from the objective function used during training (and early stopping), other aspects of model training and selection were held constant across approaches.
To give the baseline the best chance, we resampled and shuffled the training data while training the BCE models to more fully explore the space of potential updates (resulting in 150 updates instead of 3).
The best ``BCE'' and ``RBC models'' from these model sets were selected based on validation performance using \textbf{Equation \ref{eq:selection_fx}}.
We compare the resulting ``BCE" and ``RBC models'' by calculating the difference in rank-based compatibility, $\Delta \RBC$, and difference in \ac{AUROC}, $\Delta \AUROC$ (an example of this calculation can be found in \textbf{Appendix Section \ref{sec:example_replication_results}}).
We repeated this process 40 times, for every value of $\alpha$ and every value of $\beta \in \{0, 0.1, ..., 0.9, 1\}$ and compared the mean differences in both $\RBC$ and \ac{AUROC}. 

\begin{figure}[!h]
    \centering
    \includegraphics[width=\columnwidth]{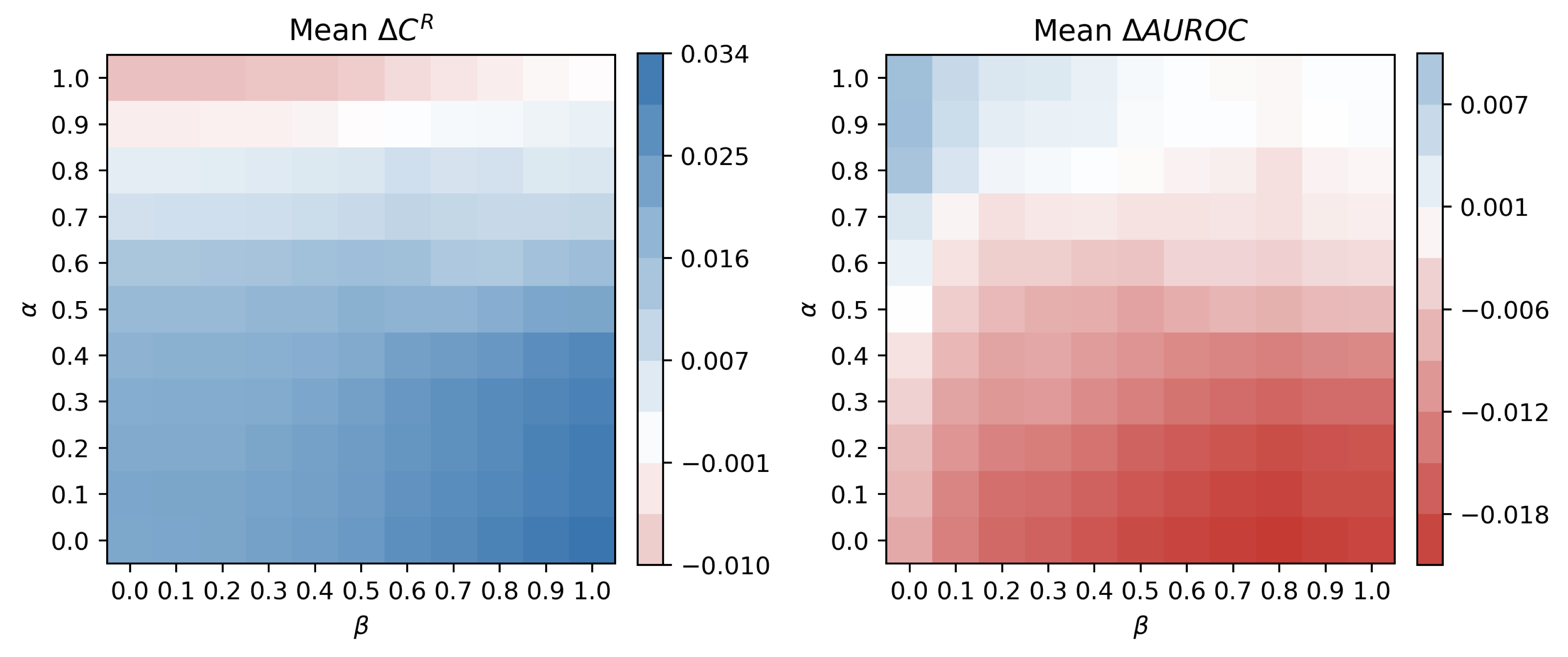}
    \caption
    {Performance Differences Between ``RBC Models'' and ``BCE Models'' With Variation of $\alpha$ and $\beta$.
    Mean value of $\Delta \RBC$ on the left and mean value of $\Delta \AUROC$ on the right, blue shows improvement of ``RBC Models'' over ``BCE models'' and red shows degradation.
    For a large majority of $\alpha$-$\beta$ pairs, there is an improvement in mean $\RBC$.
    For a smaller majority, there is a degradation in mean $\AUROC$.
    This suggests that there is a trade-off between $\AUROC$ and $\RBC$, with improved $\RBC$ coming at the cost of $\AUROC$.
    Although this trade-off exists, we note that the degradations in $\AUROC$ are often not statistically significant, while the improvements in $\RBC$ are.
    This is shown and discussed in \textbf{Appendix Section \ref{sec:improvement_statistically_significant}}.
    }
    \label{fig:selection_vs_engineering_deltas_alpha_beta}
\end{figure}

Results are displayed in \textbf{Figure \ref{fig:selection_vs_engineering_deltas_alpha_beta}}.
There is a trade-off between \ac{AUROC} and $\RBC$. For many $\alpha$-$\beta$ combinations, there is a significant gain in $\RBC$ (blue) at the cost of lower \ac{AUROC} (red) when using the proposed objective function during optimization.  
However, we note many cases in which there is a gain in compatibility without paying a penalty in terms of \ac{AUROC}.
For example, when $\alpha=0.5$ and $\beta=0.5$, we achieve a significant gain in compatibility of $\Delta \RBC=0.019$ ($95\%$ confidence interval: $0.005$, $0.035$) with an $\Delta AUROC=-0.009$ ($-0.030$, $0.011$).%
\footnote{The ``RBC models'' had the following performance: $\RBC=0.966$ ($0.948$, $0.979$) $\AUROC=0.828$ ($0.804$, $0.855$) vs. ``BCE models'' with $\RBC=0.947$ ($0.932$, $0.963$)}
By incorporating $\lrbc$ during training, it is possible to achieve improved compatibility without compromising discriminative performance. 
Out of the 121 $\alpha$-$\beta$ combinations, $57$ demonstrate statistically significant improvements in $\RBC$ while maintaining $\AUROC$; see \textbf{Appendix Section \ref{sec:improvement_statistically_significant}} for further discussion.

\begin{figure}[!h]
    \centering
    \includegraphics[width=\columnwidth]{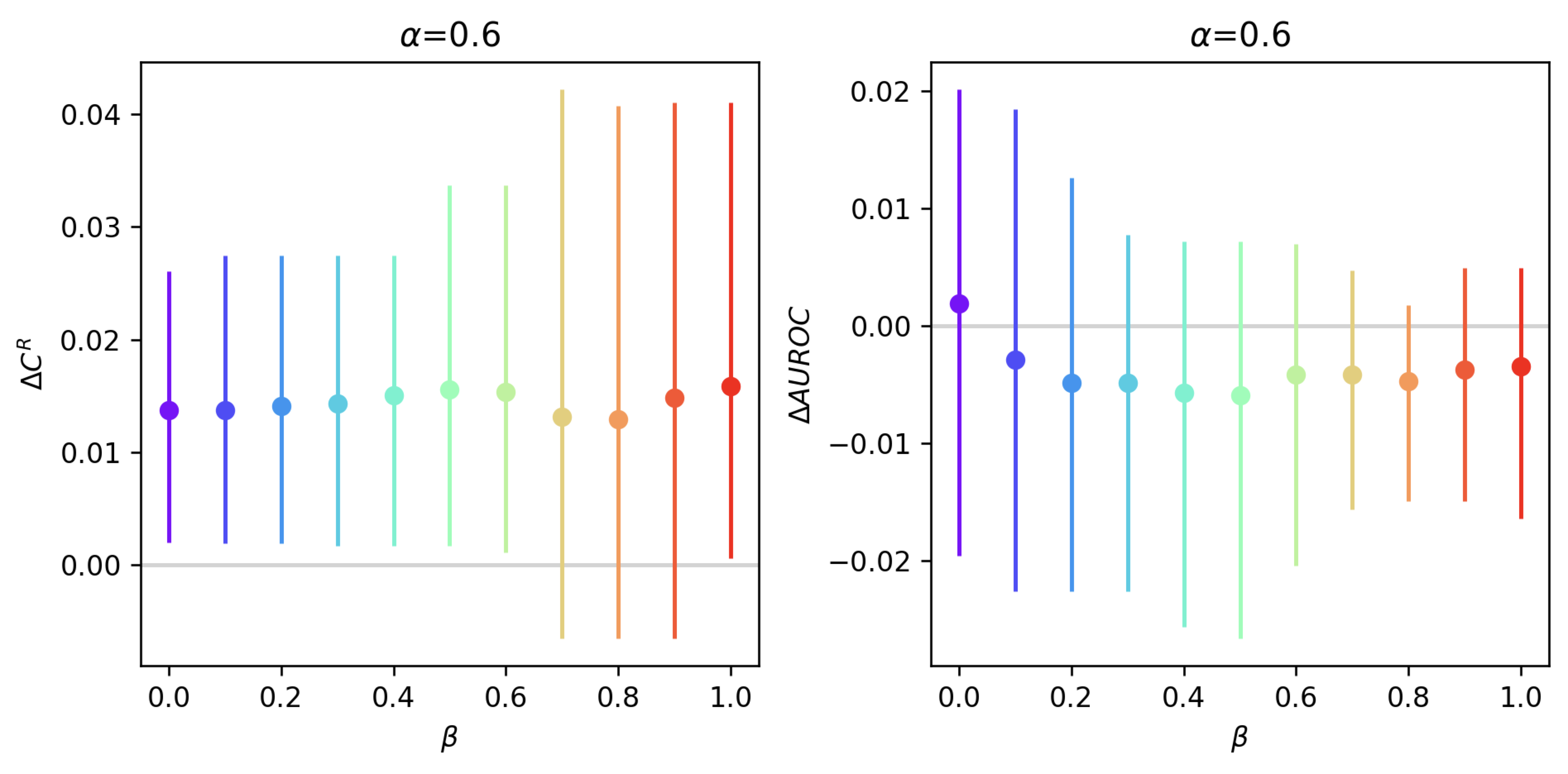}
    \caption
    {Performance Difference for $\alpha=0.6$ Sweeping Across $\beta$.
    Comparing $\Delta \RBC$ and $\Delta \AUROC$ for various $\beta$ values.
    We see that for all $\beta$ values, there is no statistically significant degradation in $\AUROC$ while for $\beta$ values less than $0.7$ we see improvement in $\RBC$.
    This suggests that ``RBC models'' yield a benefit over ``BCE models'' in this regime.
    }
    \label{fig:selection_vs_engineering_deltas_alpha_fixed_0.6}
\end{figure}

\begin{figure}[!h]
    \centering
    \includegraphics[width=\columnwidth]{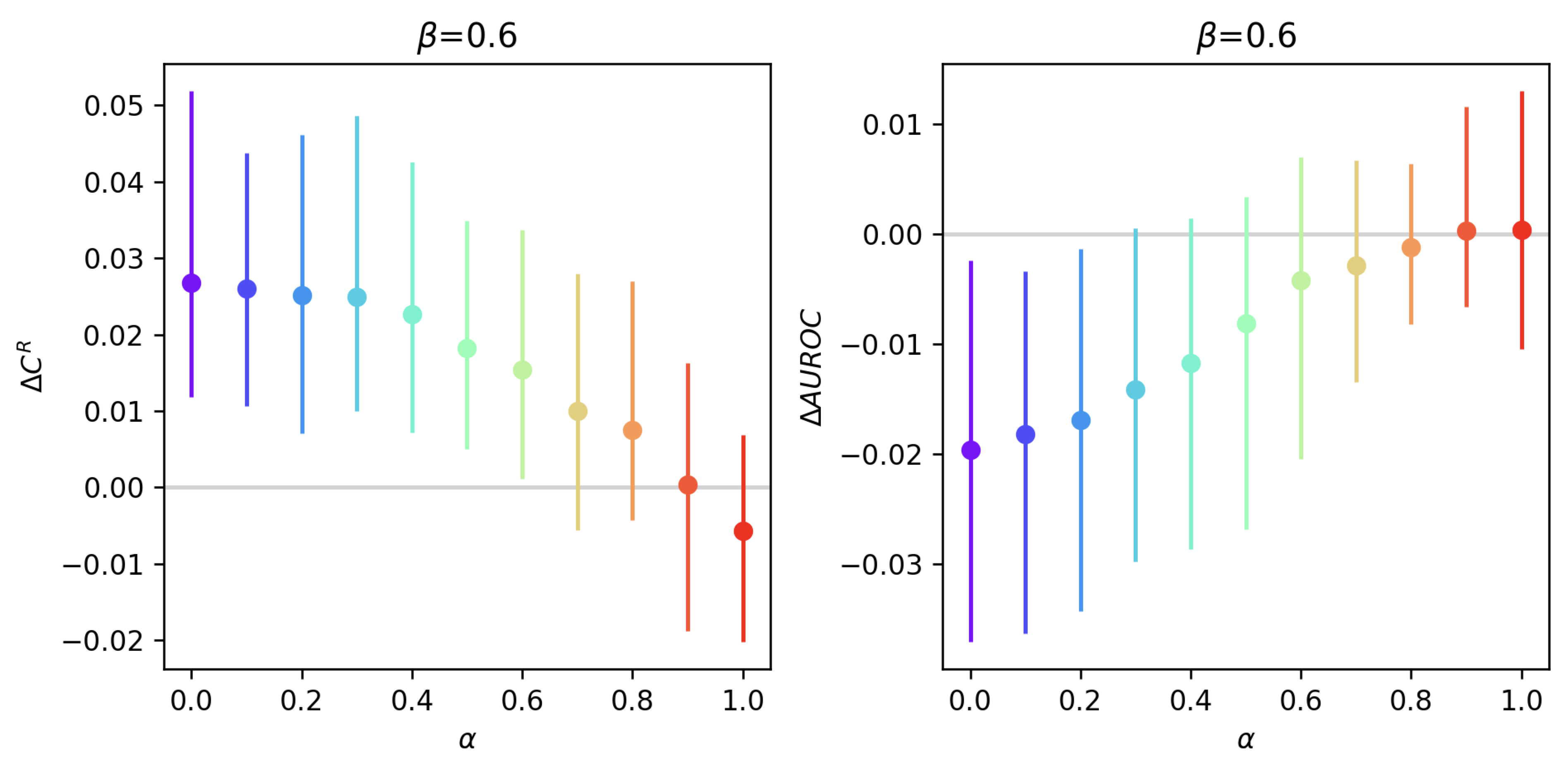}
    \caption
    {Performance Difference for $\beta=0.6$ Sweeping Across $\alpha$.
    Comparing $\Delta \RBC$ and $\Delta \AUROC$ for various $\alpha$ values.
    In this case see a more limited benefit of the ``RBC models'' over ``BCE models'', with $\alpha \in [0.3, 0.6]$ showing significant benefit in $\RBC$ and no significant degradation in $\AUROC$. 
    }
    \label{fig:selection_vs_engineering_deltas_beta_fixed_0.6}
\end{figure}

Examining results across replications for an $\alpha=0.6$ while we vary $\beta$
\textbf{Figure \ref{fig:selection_vs_engineering_deltas_alpha_fixed_0.6}},  we see that across selection options, the ``RBC model'' generally provides a better $\RBC$ (statistically significant for $\beta \leq 0.6$) without a significant decrease in \ac{AUROC} (\ie $\Delta AUROC$ is at or close to zero). In \textbf{Figure \ref{fig:selection_vs_engineering_deltas_beta_fixed_0.6}}, we set $\beta=0.6$ during the selection process for both the ``RBC models'' and the ``BCE models'', and sweep $\alpha$ during training ``BCE models''.
Again, we observe that for specific $\alpha$ values (e.g., $\alpha=0.3-0.6$), we can significantly improve compatibility without penalizing \ac{AUROC} performance.

\textit{These empirical results suggest that by incorporating rank-based compatibility into the objective function during training, we can generate model updates with larger $\RBC$ values than obtained through standard update generation procedures (\ie minimizing for $\lbce$ alone).}
Moreover, while there is often a trade-off between $\RBC$ and \ac{AUROC}, achieving gains in $\RBC$ while maintaining $\AUROC(f^u)$ is possible.

\section{Discussion \& Conclusion} 

When selecting among potential updated clinical risk stratification models, it may be important to consider compatibility with existing models already in use.
In this study, we propose the first rank-based compatibility measure, $\RBC$, which measures the concordance in ranking between two models. 
We illustrate the connection between $\RBC$ and discriminative model performance.
This relationship suggests that increased rank-based compatibility accompanies improved discriminative performance, as the lower bound of rank-based compatibility increases as each model's discriminative performance increases.
Despite this relationship, we show empirically that it is improbable to observe very high levels of rank-based compatibility through standard updated model development, which tends to focus on optimizing discriminative performance.
These findings motivate methods that enable developers to build models with good discriminative performance and rank-based compatibility.
As such, we introduce a new differentiable rank-based incompatibility loss function that can be used when training updated models to further optimize for rank-based compatibility.

We used the MIMIC-III dataset to compare our proposed approach to generating model updates to a standard approach that optimizes for binary cross-entropy alone. 
Our results highlight standard updated model development's limitations in identifying model updates with very high compatibility.
Using our proposed approach, we identify models with equivalent discriminative performance yet significantly better compatibility. 
However, if rank-based compatibility is greatly emphasized over discriminative performance, then improvements may come at a cost.

The rank-based compatibility measure serves a different role than the original backwards trust compatibility measure proposed by \citet{RN512}.
Depending on the use case, one may choose one over the other. 
Use cases that strongly depend on decision thresholds, such as sending a notification when a patient risk estimate exceeds a specific threshold, may correspond to clinician mental models best represented by $\BTC$.
In settings where the decision may depend on the state of the system, such as hospital admission decisions, which are impacted by the number of patients in the emergency department \citep{RN1326}, the $\RBC$ may better represent clinician mental models because it is not tied to a fixed threshold.
Additionally, the complexity of this evaluation grows proportionally with the number of models and thresholds being considered.
Thus, if there are many potential thresholds, it may be more effective to use $\RBC$ directly.



Although we know the absolute scale of rank-based compatibility with $0$ denoting ``no compatibility'' and $1$ denoting ``perfect compatibility'', we do not have a sense of what the numbers in between mean and how they compare across model updates.
Ideally, we would like to have a sense of what is an excellent rank-based compatibility value, like we do with the \ac{AUROC} measure (\eg $\AUROC(f) > 0.85)$.
This will likely come with further study of models being updated across different tasks.
One advantage $\RBC$ does present is that its improvements can be directly compared against improvements in \ac{AUROC} by examining the \ac{POP} variables.

While we discussed the different use cases for $\RBC$ vs. $\BTC$, we did not explore users' preferences.
Although there may be update tasks for which the $\RBC$ measure is better suited, we have not yet characterized the relationship between rank-based compatibility and user mental models.
For example, a sepsis detection system that flags patients as at risk \citep{RN1392} or sends an alert notification \citep{RN982} may be a good candidate for the $\BTC$ compatibility measure.
Users in these cases would expect consistent correct classification of patients when the underlying model is updated.
If users interact with the model to help risk stratify their patients, then the $\RBC$ measure may be a better choice.
Tools used for cardiovascular event risk stratification \citep{RN1393} and in-hospital deterioration risk stratification \citep{RN1394, RN1328} may be more effectively updated using $\RBC$.

Like backwards trust compatibility, rank-based compatibility captures the ``global'' user perspective.
Modifying rank-based compatibility to focus on individual user perspectives may lead to better compatibility and parity with user expectations \citep{RN764, RN1337}.
We have focused our study of rank-based compatibility exclusively on when the updated model continues the correct behaviors established by the original model.
Previous user studies have shown that user mental models are influenced by the error behavior of classification models \citep{RN514}.
This may hold for risk stratification models, motivating the study of incorrect ranking in conjunction with rank-based compatibility.
We believe there is much work to do with this measure in terms of human user studies.

Finally, the primary analysis we present is based on the experimental setup developed by \citet{RN512}.
Although this was intentionally done to enable the comparison of the $\BTC$ and $\RBC$ it is not an exhaustive evaluation.
Notably, future work may benefit from the exploration of different tasks, datasets, and model architectures.
Some tasks like survival analysis \citep{RN1374} may be able to use the general form of $\RBC$, \textbf{Equation \ref{eq:gf_rbc}}.
Different model architectures may need adaption of the joint optimization of performance and compatibility proposed by this work.
Additionally, there are real world complexities that are unaccounted for in this analysis, such as outcome censoring due to clinician interventions based on model predictions \citep{RN705} and the impact of deployment infrastructure changing as models are updated \citep{RN1048}.

These limitations notwithstanding, the new rank-based compatibility measure and incompatibility loss present a novel way to think about model maintenance and updating models, beyond simply optimizing for \ac{AUROC}.
Furthermore, optimizing the rank concordance between the output of two models, rather than thresholded predictions, may be more robust to calibration shifts, a commonly observed phenomenon in healthcare \citep{RN873, RN841, RN874}.
We expect this new measure applies in evaluating healthcare risk stratification models.
However, there are likely settings in domains beyond healthcare that would similarly benefit from such rank-based measures.
Overall, this work enables the evaluation and development of model updates that have the potential to lead to better clinician-model team performance.

\acks{E\"{O} was supported by NIH grant T32GM007863 and JW was supported by the Alfred P. Sloan Foundation. 
The authors would like to thank the anonymous reviewers and editors of the 2023 Machine Learning for Healthcare Conference for their thoughtful feedback. 
}

\bibliography{bibliography/20220815}

\newpage
\appendix

\section{Background}

\subsection{Decision Threshold Dependence of $\BTC$}
\label{sec:decision_threshold_dependence_of_btc}

Like accuracy, $\BTC$ is highly dependent on model thresholds.
We illustrate this in \textbf{Figure \ref{fig:btc_various_thresholds}}, which shows the dependence of $\BTC$ on the updated model decision threshold ($\tau^u$).
Depending on the choice of $\tau^u$ $\BTC(f^o, f^u)$ may be $\tfrac{1}{2}$, $\tfrac{3}{4}$, or $1$.
Because every patient-pair is correctly ordered by both models in this example, the $\RBC$ equals $1$.
If $\tau^o$ had been set to a much larger (or smaller) value, then it would have been possible for $\BTC$ values of $0$ to occur for this example.
Ultimately, poorly chosen decision thresholds or models miscalibrated with one another may demonstrate poor $\BTC$ even if both the original and updated models have good discrimination and concordance in their correct patient-pair rankings ($\RBC$).

\section{Methods}

\subsection{General Form Rank-Based Compatibility}

\textbf{Equation \ref{eq:rbc}} defines rank-based compatibility for risk stratification models operating over binary labels.
Rank-based compatibility is not limited to use only in situations where the outcomes are binary.
The core concept can be applied to any set of patient labels that can be ordered (\eg integer or real values).
We now present a general form rank-based compatibility equation that can be employed in these situations. 

\begin{equation}
    \label{eq:gf_rbc}
        \RBC(f^o, f^u) :=
        \frac{
            \sum \limits_{i \in I}
                \sum \limits_{j \in I}
                    \ind(\hat{p}_i^o < \hat{p}_j^o)
                    \cdot
                    \ind(\hat{p}_i^u < \hat{p}_j^u)
                    \cdot
                    \ind(y_i < y_j)
            }{
            \sum \limits_{i \in I}
                \sum \limits_{j \in I}
                    \ind(\hat{p}_i^o < \hat{p}_j^o)
                    \cdot
                    \ind(y_i < y_j)
        } 
\end{equation}

This equation has several minor changes from \textbf{Equation \ref{eq:rbc}}.
Differences in the summation indices enable the equation to evaluate every patient-pair and an additional term ($\ind(\hat{p}_i^o < \hat{p}_j^o)$) in the numerator and denominator checks if this patient-pair is ordered correctly by the label.

\subsection{\ac{POP} Variable Bounds}
\label{sec:app_pop_variable_bounds}

Given the assumption that the updated model is at least as good as the original model and that both models are better than random (\ie $0.5 \leq AUROC(f^o) \leq AUROC(f^u)$) and the relationships established in \textbf{Table \ref{tab:pop_table}}, several constraints exist on the \ac{POP} variables.
These are:
\begin{equation*}
    \begin{split}
        \AUROC(f^o)+\AUROC(f^u)-1  \leq \phi^{++} \leq \AUROC(f^o)
        \\
        0 \leq \phi^{+-} \leq 1-\AUROC(f^u) 
        \\
        0 \leq \phi^{-+} \leq 1-\AUROC(f^o) 
        \\
        0 \leq \phi^{--} \leq 1-\AUROC(f^u)
    \end{split}
\end{equation*}

\noindent
In this study, we focus only on the \ac{POP} variable that represents both models ranking patient-pairs correctly, $\phi^{++}$, as it is the only one used directly in $\RBC$.
So we briefly discuss how we derive its bounds.
The minimum value $\phi^{++}$ can take is the smallest proportion of correctly ordered patient-pairs by both models.
Since the \acp{AUROC} of both models must be at least $0.5$, the smallest this proportion is when there is minimal overlap in the set of correctly ordered patient-pairs for each model.
This is the sum of the two \acp{AUROC} subtracted by 1.
The maximal value for $\phi^{++}$ is determined by the smaller of the two model’s AUROC which is $\AUROC(f^o)$.

\subsection{Lower-bound of $\RBC$}
\label{sec:app_lower_bound_of_rbc}

\begin{figure}[!ht]
    \centering
    \includegraphics[width=0.75\columnwidth]{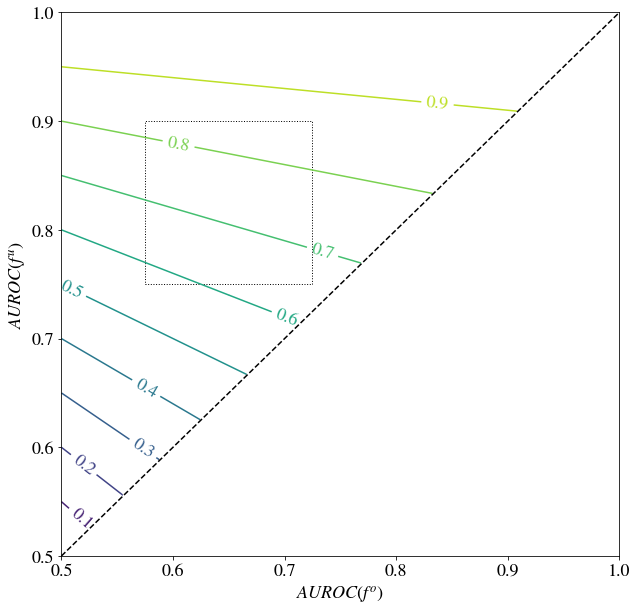}
    \caption[Lower bound of Rank-Based Compatibility]
    {Lower bound of $\RBC$ with respect to the \ac{AUROC} of the original and updated models.
    The lower bound increases as both models' performance increases.
    The boxed region with dotted lines demarcates a typical discriminative performance region.
    In this region we would expect to observe $\RBC$s no smaller than $0.5$.
    }
    \label{fig:rank_based_compatibility_lower_bound}
\end{figure}

We produce a plot for the lower bound of the rank-based compatibility measure (\textbf{Figure \ref{fig:rank_based_compatibility_lower_bound}}).
For the regime of model updating that we are interested in $0.5 < \AUROC(f^o) \leq \AUROC(f^u) \leq 1$), only the lower bound of $\RBC$ varies, increasing as the discriminative performance of the two models grows.

\subsection{Central Tendency of $\RBC$}

Though the bounds suggest that higher compatibility is partially correlated with higher discriminative performance, we expect that in practice updated models $\RBC$ values will have a central tendency.
We present a brief analytical sketch to explore this behavior. 
Note, we do not seek to create a distribution for the $\RBC$ generally; instead, we seek to build intuition for how $\RBC$ may vary with both models' \ac{AUROC}.
This analytical approach is based on a combinatorial argument.
We analyze the number of ways a given $\RBC$ can occur given AUROCs for the original and updated models.
This analysis is based on how each model ranks each patient-pair.
A patient-pair's ranking for a given model is whether that model correctly ranks (\eg $\hat{p}_i<\hat{p}_j$ for the updated model) or incorrectly ranks that patient pair.

We can use the ranking of all patient-pairs to represent the behavior of original and updated models.
All patient-pairs are distributed between two sets: correctly and incorrectly ranked.
Suppose we constrain the distribution of patient-pairs between these two sets to align with the discriminative performance of the model being represented.
In that case, we can then get a sense of the number of patient-pairs that both models rank correctly.
This number is $m^{++}$ and can be directly used to calculate the $\RBC$ as per \textbf{Equation \ref{eq:rbc}}.
As mentioned in \textbf{Appendix Section \ref{sec:app_lower_bound_of_rbc}}, $m^{++}$ may range between $m^{o+} + m^{u+} - m$ and $m^{o+}$, corresponding to the bounds $\RBC$ introduced in \textbf{Equation \ref{eq:rbc_bounds}}.
Assuming models do not have any restrictions on how patient-pairs may be ranked, we count the number of ways that each value of $m^{++}=k$ can be achieved given that each model meets a specific \ac{AUROC}.
We refer to this count as $\nu$, where $\nu = |\{m^{++}=k | m^{o+}, m^{u+}\}| $.

$\nu$ is the numerator of the hypergeometric distribution with parameters related to the number of patient-pairs correctly ranked by the original and updated models.
The number of patient-pairs that both models ranked correctly, $m^{++}=k$, is defined in relation to the number of total patient-pairs, $m$, the number of patient-pairs we are interested in selecting, $m^{o+}$, and the number of selections, $m^{u+}$.
The number of combinations that produce a given $m^{++}=k$ is as follows:
\begin{equation*}
    \label{eq:compatibility_combination_counts}
    \nu
    = 
    \binom{m^{o+}}{k} \binom{m-m^{o+}}{m^{u+}-k}
\end{equation*}

\noindent
The location and shape of this function provide us with a sense of the behavior conditional on the two model's \ac{AUROC}.
In \textbf{Appendix Section \ref{sec:app_central_tendency_of_rbc}} we plot this function and show where we would expect its maxima to occur.
From this analysis, we expect $\RBC$ to be centered around the \ac{AUROC} of the updated model and should have a strong central tendency behavior.

While we do not believe this specific center to hold for all data generating processes and model updating procedures, we hypothesize that the central tendency of $\RBC$ does.
In \textbf{Section \ref{sec:chpt4_ear_rbc_central_tendency}}, we investigate the central tendency of $\RBC$ for original and updated models trained using real data.
The above analysis is still illuminating as it provides a way to estimate the relative number of combinations between different rank-based compatibility levels.
There are many more ways for an updated model to achieve moderate rank-based compatibility (near the value of the AUROC of the updated model) than a very high level of compatibility (\eg above 0.95).
This suggests that achieving high rank-based compatibility may only be possible with directed search efforts.

\subsection{Maxima of Central Tendency of $\RBC$}
\label{sec:app_central_tendency_of_rbc}

The location of the maxima and shape of this function provides us with a sense of the behavior of $\RBC$ conditional on maintaining a fixed level of discrimination.
We would expect this function's maxima to coincide with the mode of the corresponding hypergeometric distribution.
For large values of $m^{o+}$, $m^{u+}$, and $m$ we expect the mode of the hypergeometric distribution to be approximately equal to its mean.
\textbf{Equation \ref{eq:compatibility_combination_counts}} has its maxima at $m^{++}=k^*$, where $k^*$ is the value that provides the largest number of combinations.%
\footnote{
This maxima is expressed in terms of $m^{++}$, which can be converted to be in terms of $\RBC$ by dividing by $m^{o+}$.
This maxima occurs at $\frac{m^{o+}}{m^{o+}}\frac{m^{u+}}{m} = \frac{m^{u+}}{m} = \AUROC(f^u)$.
}%
This is:

\begin{equation*}
    \begin{split}
    k^* & =
    \Bigg\lfloor \frac{(m^{o+}+1)(m^{u+}+1)}{m+2} \Bigg\rfloor 
    \\
    & \approx 
    \frac{m^{o+} m^{u+}}{m} \text{ for large $m^{o+}$, $m^{u+}$, and $m$}.
    \end{split}
\end{equation*}

\begin{figure}[ht]
    \centering
    \includegraphics[width=\columnwidth]{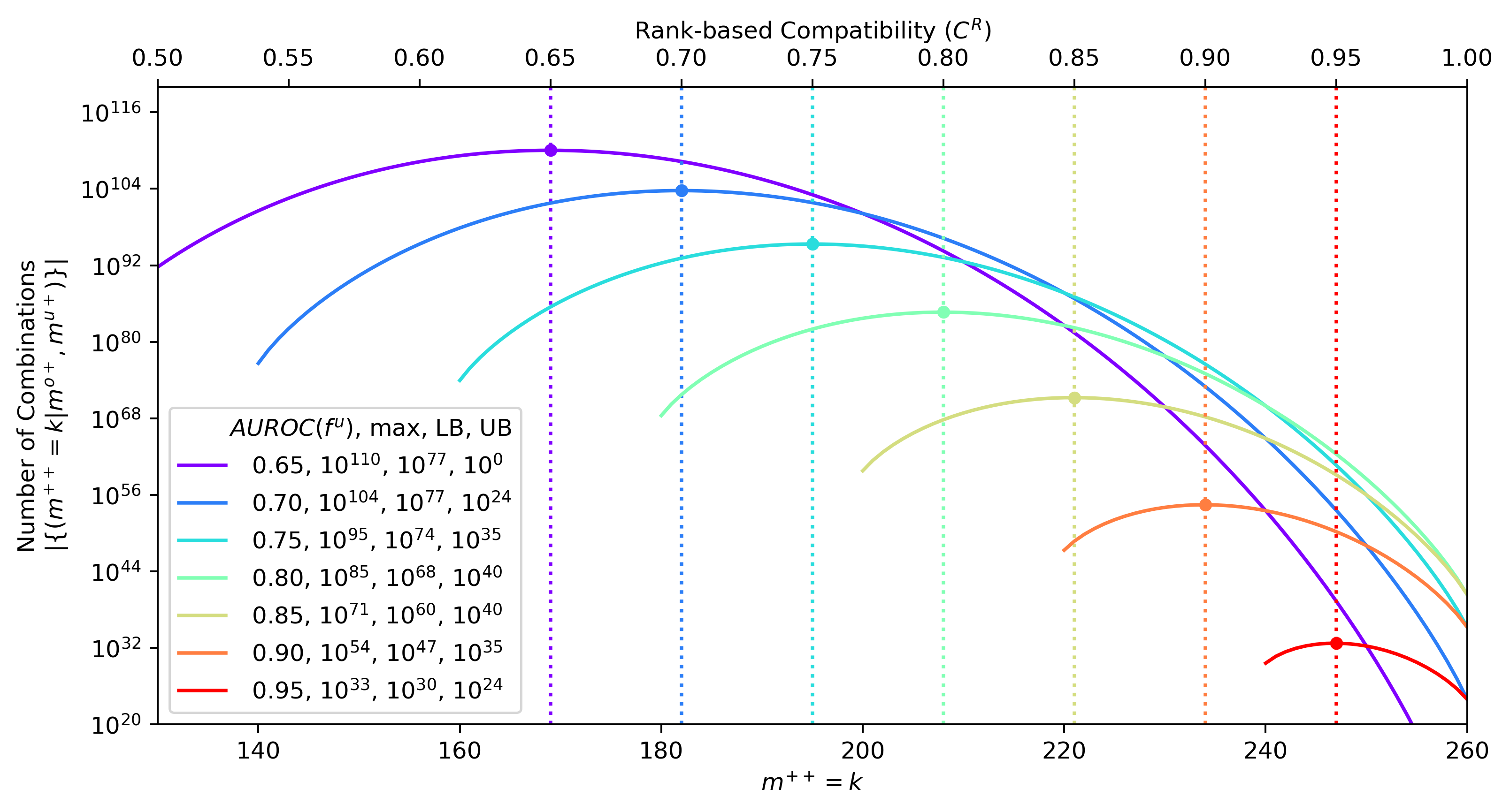}
    \caption[Central Tendency of $\RBC$]
    {
    Number of combinations that yield a given $\RBC$ value for an original-updated model pair.
    All model pairs have $\AUROC(f^o)=0.65$ and $\AUROC \in [0.65, 0.95]$ ($m=400$).
    The updated model's \ac{AUROC} is plotted as a vertical dotted line.
    The $m^{++}=k^*$ value that achieves the largest number of combinations is plotted as a dot on the curves.
    This point aligns with $\AUROC(f^o)$.
    These curves exhibit a strong central tendency as the number of combinations decreases increasingly (note the logarithmic vertical axis) as $m^{++}=k$ diverges from the $k^*$.
    }
    \label{fig:compatibility_central_tendency}
\end{figure}

We can then plot \textbf{Equation \ref{eq:compatibility_combination_counts}} to investigate the behavior of $\RBC$ given $\AUROC(f^o)$ and $\AUROC(f^u)$.
\textbf{Figure \ref{fig:compatibility_central_tendency}} shows the number of combinations for each $\RBC$ value given original-updated model pairs.
Each model pair had the same original model performance ($\AUROC(f^o) = 0.65$), and the updated performance ranged between ($\AUROC(f^u) \in [0.65, 0.95]$).
Examination of these curves reveals several findings.
First, the $k^*$ for each model pair aligns with the \ac{AUROC} of the updated model.
Second, these curves exhibit a strong central tendency as the number of combinations decreases exponentially  (note the logarithmic vertical axis) as $m^{++}=k$ diverges from $k^*$.

\newpage
\section{Experiments \& Results}

\subsection{Computing Environment}

This analysis was conducted using Python on a server running Ubuntu 16.04.07 with 112 x86 CPU cores and 503GB of RAM.
Experiments and analyses were run using Python version 3.7.4.

\subsection{Example Replication Results}
\label{sec:example_replication_results}

\begin{figure}[!ht]
    \centering
    \includegraphics[width=\columnwidth]{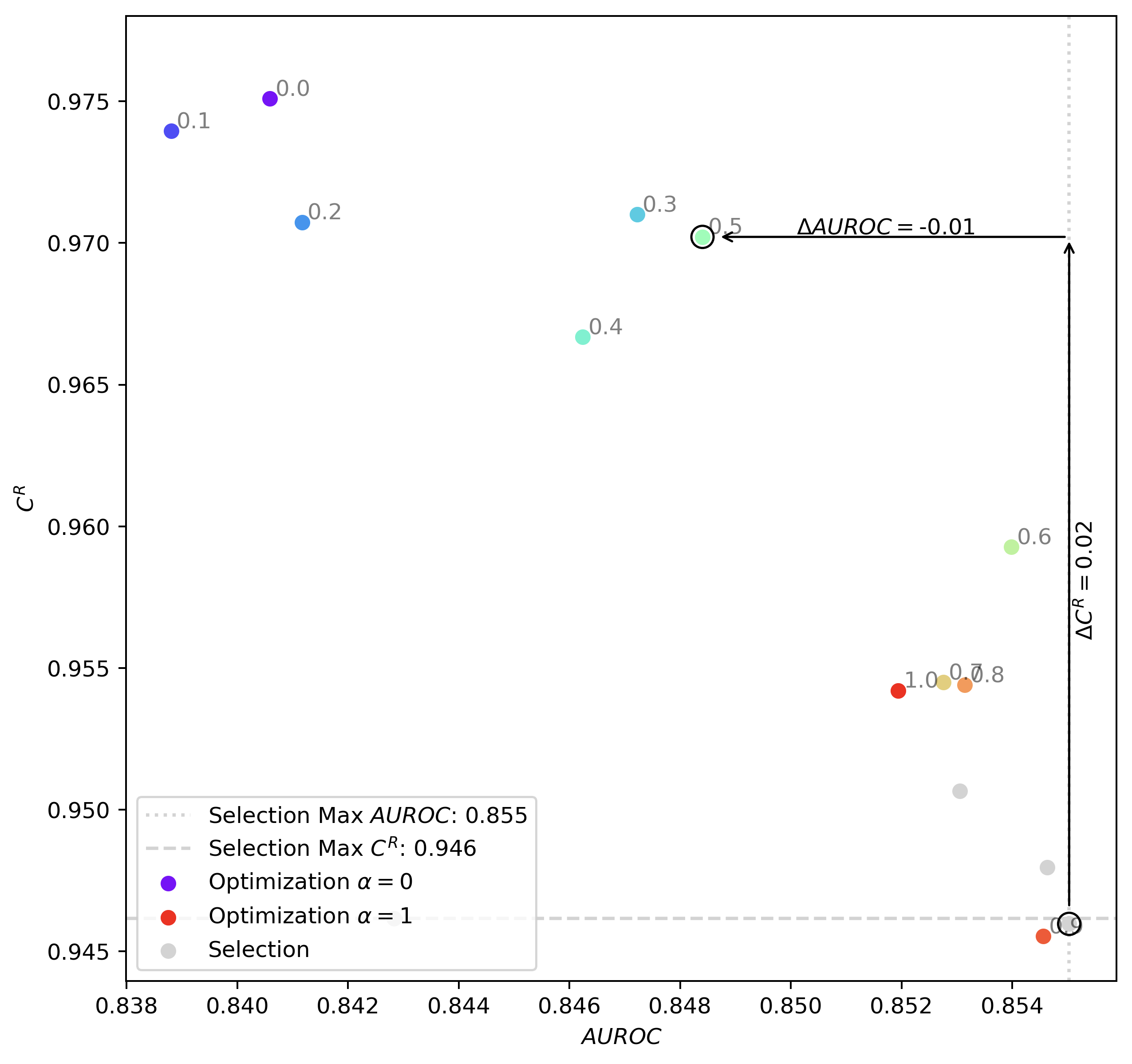}
    \caption[Example of Engineered Model vs. Selection Model Results]
    {Example of Engineered Model vs. Selection Model Results.
    The \ac{AUROC} and $\RBC$ calculated on held-out evaluation dataset are reported for the engineered models and a subset of the selection models.
    In this example, we note that the circled engineered model ($\alpha=0.5$) induces a positive $\Delta \RBC$, which denotes an increase in $\RBC$, and a negative $\Delta \AUROC$ which indicates a reduction in \ac{AUROC}.
    }
    \label{fig:example_replication_selection_vs_engineering}
\end{figure}

In \textbf{Figure \ref{fig:example_replication_selection_vs_engineering}} we show the $\RBC$ and \ac{AUROC} values calculated on the held-out evaluation data for all of the engineered models and a subset of the selection models.
This subset represents the selection models along the pareto frontier of the trade-off between $\RBC$ and \ac{AUROC} (calculated using the updated model validation data).
We also depict how $\Delta \RBC$ and $\Delta \AUROC$ would be calculated between the engineered model where $\alpha=0.6$ and the selected candidate update with the best \ac{AUROC}.

For this example, we note that the circled engineered model induces a positive $\Delta \RBC$, which denotes an increase in $\RBC$, and a negative $\Delta \AUROC$, which represents a reduction in \ac{AUROC}.
Although the $\Delta \AUROC$ is negative, this does not mean that this updated model performs worse than the original model, which has an $\AUROC=0.805$.
Instead, the engineered update ($\AUROC=0.848$) does not perform as well as the best-performing selection model ($\AUROC=0.855$).

\subsection{Improvement in $\RBC$ Compared with Distribution From Standard Model Updating}

\begin{figure}[!ht]
    \centering
    \includegraphics[width=\columnwidth]{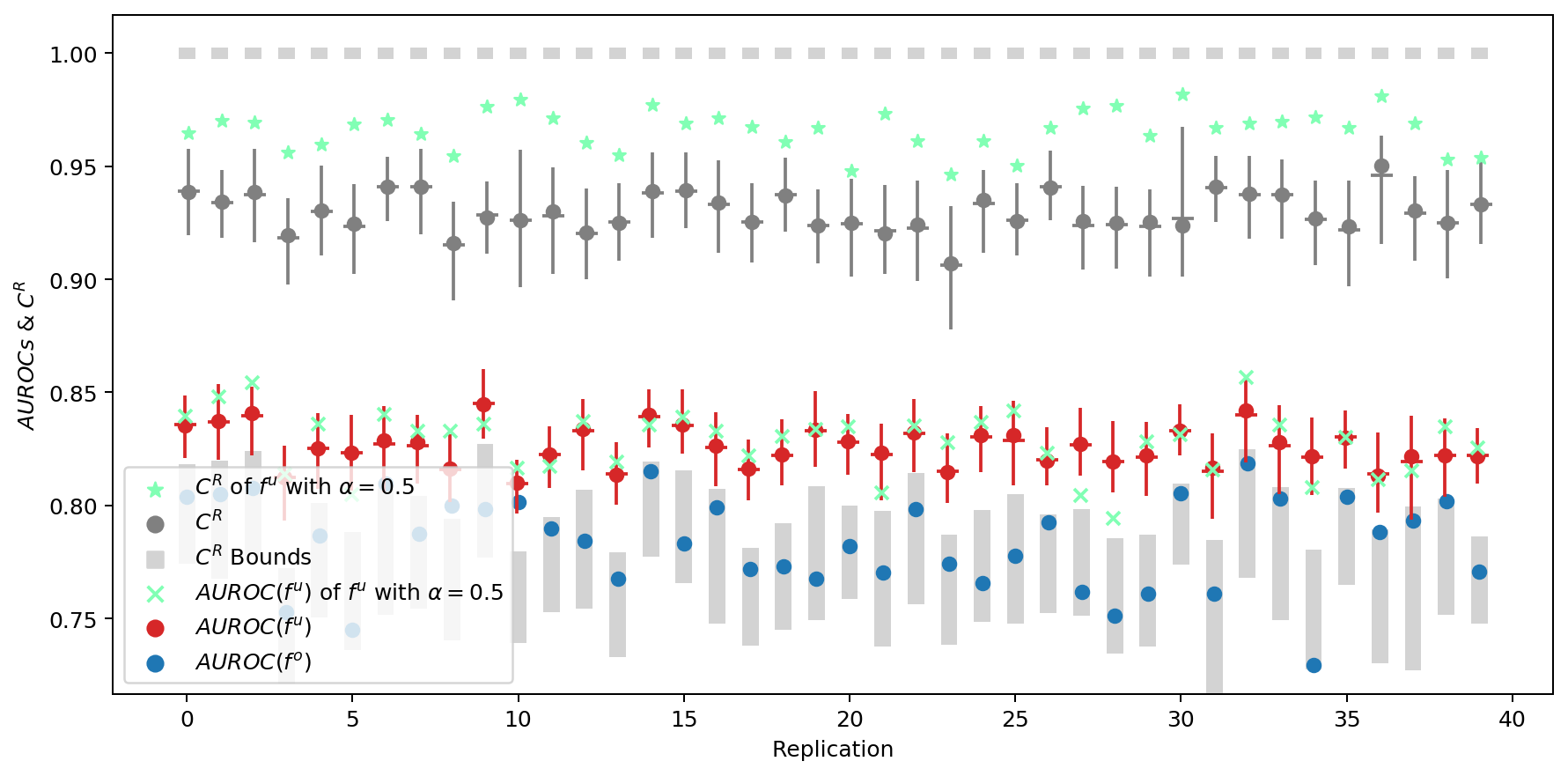}
    \caption[Improvement in $\RBC$ of ``RBC Models'' at $\alpha=0.5$]
    {Improvement in $\RBC$ of ``RBC Models'' at $\alpha=0.5$.
    We note how for nearly all of the replications the ``RBC Model'' produces $\RBC$ values exceeding those produced by the ``BCE Models''.
    The $\AUROC$ values are relatively in line with one another.
    }
    \label{fig:mortality_update_compatibility_with_example_optimization}
\end{figure}

We show that ``RBC Models'' can produce $\RBC$ values greater than what is observed through standard model updating procedures with little cost in terms of $\AUROC$. See \textbf{Figure \ref{fig:mortality_update_compatibility_with_example_optimization}}.

\subsection{$\phi^{++}$ Central Tendency}
\label{sec:phi_plus_plus_central_tendency}

\begin{figure}[!h]
    \centering
    \includegraphics[width=\columnwidth]{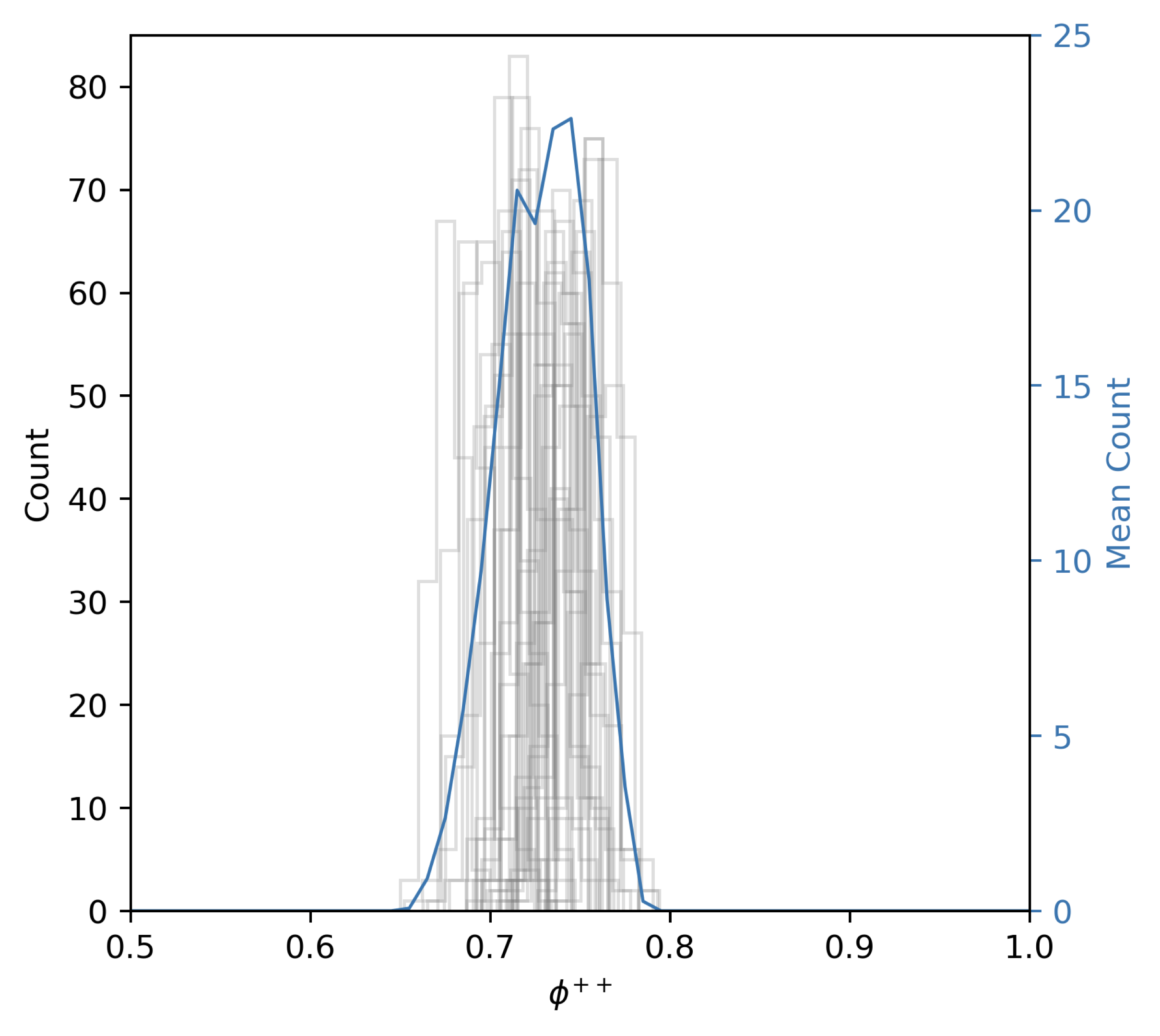}
    \caption
    {Central Tendency of $\phi^{++}$.
    $\phi^{++}$ for each replication plotted in gray, bin size=$0.01$.
    Note a small amount of uniform jitter was added during plotting.
    The mean of these histograms across all replications is plotted in blue.
    }
    \label{fig:central_tendency_of_phi_plus_plus}
\end{figure}

As mentioned above, the distributions of $\RBC$ shown in \textbf{Figure \ref{fig:compatibility_central_tendency}} shift in relation to the \acp{AUROC} of the models.
To control for this shift we examined the \ac{POP} variable $\phi^{++}$.
We did this by calculating the $\phi^{++}$ for each updated model.
We then created a histogram for all updated models (histogram bin size=$0.01$).
This procedure was repeated for all $40$ replications.
We then averaged the bin counts over all the replications.
These results are plotted in \textbf{Figure \ref{fig:central_tendency_of_phi_plus_plus}}

From this plot, we see that each replication has a strongly peaked histogram and that the mean distribution of $\phi^{++}$ has a robust central tendency.

\subsection{Improvement of $\Delta \RBC$ and Non-Degradation of $\Delta \AUROC$}
\label{sec:improvement_statistically_significant}

\begin{figure}[!h]
    \centering
    \includegraphics[width=\columnwidth]{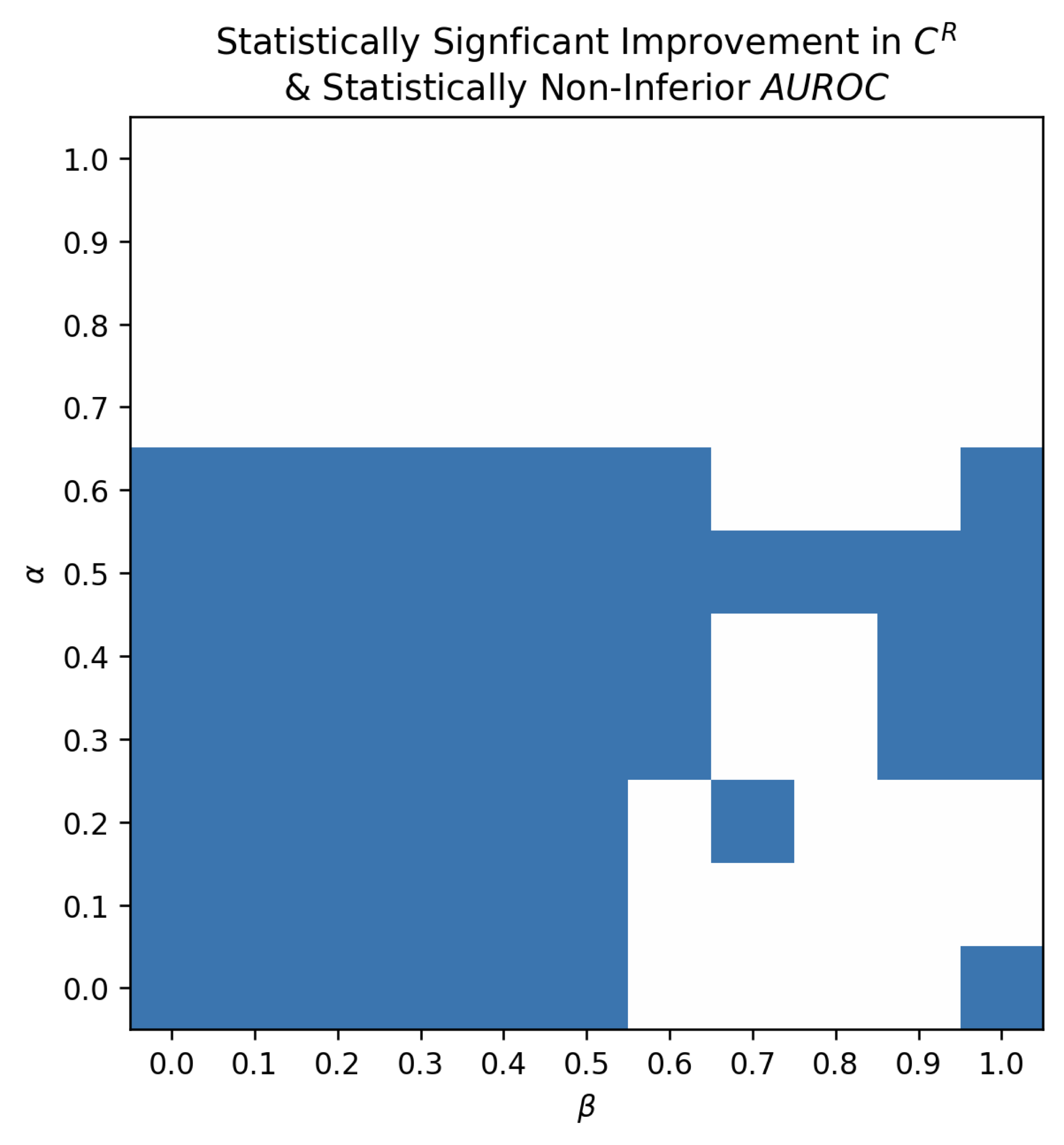}
    \caption
    {$\alpha$-$\beta$ Combinations Showing Improvement.
    All $\alpha$-$\beta$ combinations with a statistically significant improvement in $\RBC$ without a statistically significant degradation in \ac{AUROC} are depicted in blue.
    }
    \label{fig:selection_vs_engineering_improvement_statistically_significant}
\end{figure}

The graphs presented in \textbf{Section \ref{sec:chpt4_ear_engineered_loss_vs_standard_selection}} show the mean $\Delta \RBC$ and $\Delta \AUROC$.
However, we examined the $95\%$ confidence intervals to assess statistically significant differences.
We determined if there was a statistically significant improvement in $\Delta \RBC$ (\ie the confidence interval does not include $0$) and if there was not a statistically significant degradation in $\Delta \AUROC$ (\ie the confidence interval does include $0$).
The $\alpha$-$\beta$ combinations that met these criteria are in blue in \textbf{Figure \ref{fig:selection_vs_engineering_improvement_statistically_significant}}.
We note that $57$ out of the $121$ $\alpha$-$\beta$ combinations show an improvement with the inclusion of the $\lrbc$ loss.

\begin{figure}[!h]
    \centering
    \includegraphics[width=\columnwidth]{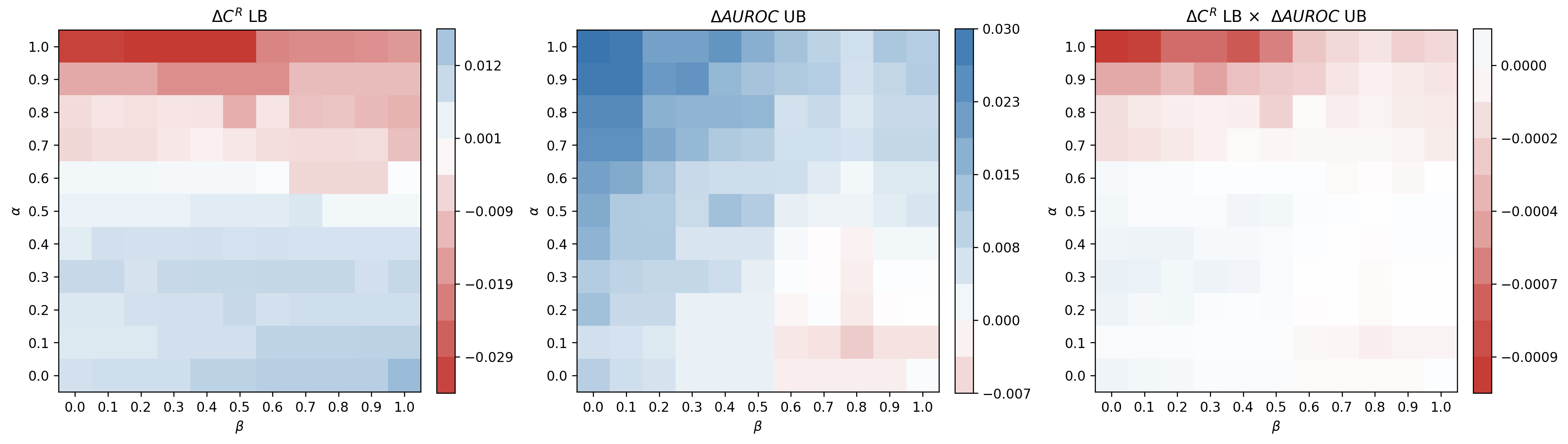}
    \caption
    {Details of $\alpha$-$\beta$ Combinations Showing Improvement.
    In the left panel, we show the $95\%$ confidence interval lower bound for $\Delta \RBC$.
    The blue areas represent $\alpha$-$\beta$ combinations that yield a statistically significant improvement (\ie $\Delta \RBC > 0$). 
    In the middle panel, we show the $95\%$ confidence interval upper bound for $\Delta \AUROC$.
    The blue areas represent combinations without statistically significant degradation (\ie $\Delta \AUROC \geq 0 $).
    In the right panel, we show the product of the two previous panels to show how we arrived at the above results.
    }
    \label{fig:selection_vs_engineering_improvement_breakdown}
\end{figure}

In order to characterize the $\alpha$-$\beta$ combinations where we see this improvement, we plot the critical confidence interval values (the lower bound of $\Delta \RBC$ and the upper bound of $\Delta \AUROC$) along with their product in \textbf{Figure \ref{fig:selection_vs_engineering_improvement_breakdown}}.

From \textbf{Figure \ref{fig:selection_vs_engineering_improvement_breakdown}}, we can see there are several ``regions'' of $\alpha$-$\beta$ combinations.
When $\alpha$ is high (\ie $\alpha \geq 0.7$), then we observe that we may not have a statistically significant improvement in $\Delta \AUROC$ (the red region on top of the left panel).
This makes sense.
As $\alpha$ increases, we de-emphasize the importance of $\RBC$, and thus, there should be little difference (in terms of $\RBC$) between the ``RBC'' and ``BCE models''.
We note that when $\alpha$ is low, and $\beta$ is high (\ie $\alpha \leq 0.4$ and $\beta \geq 0.6$) that we may have a statistically significant degradation in $\Delta \AUROC$.
Again, this makes sense, as this $\alpha$-$\beta$ combination represents training ``RBC models'' to focus on $\RBC$ but then attempting to select models based on \ac{AUROC}.
This training-selection discrepancy would disadvantage the ``RBC models'' in terms of \ac{AUROC}.
Thus, when we overlay these areas of interest, we see that we generally tend to observe statistically significant improvements in $\Delta \RBC$ that come without an \ac{AUROC} cost in the region of low $\alpha$ and low $\beta$ (\ie $\alpha \leq 0.6$ and $\beta \leq 0.5$).
Notably, this region aligns with model developers seeking to emphasize compatibility as a part of their updated model development process.

\subsection{$\BTC$ Across Thresholds}
\label{sec:btc_across_thresholds}

As discussed in \textbf{Section \ref{sec:backwards_trust_compatibility}}, $\BTC$ depends on setting a decision threshold for each model in the model-pair.
To help contextualize how various thresholds impact $\BTC$ we conduct an additional analysis of the main experiment discussed in \textbf{Section \ref{sec:chpt4_ear_engineered_loss_vs_standard_selection}}.
In this experiment, we sweep the thresholds for the original model, $\tau^o$, and the updated model, $\tau^u$, and find the maximum achievable $\BTC$ over all of the ``BCE models''.
This analysis can be used to observe $\BTC$ across multiple thresholds, as per \citet{RN1335}.

\begin{figure}[!h]
    \centering
    \includegraphics[width=\columnwidth]{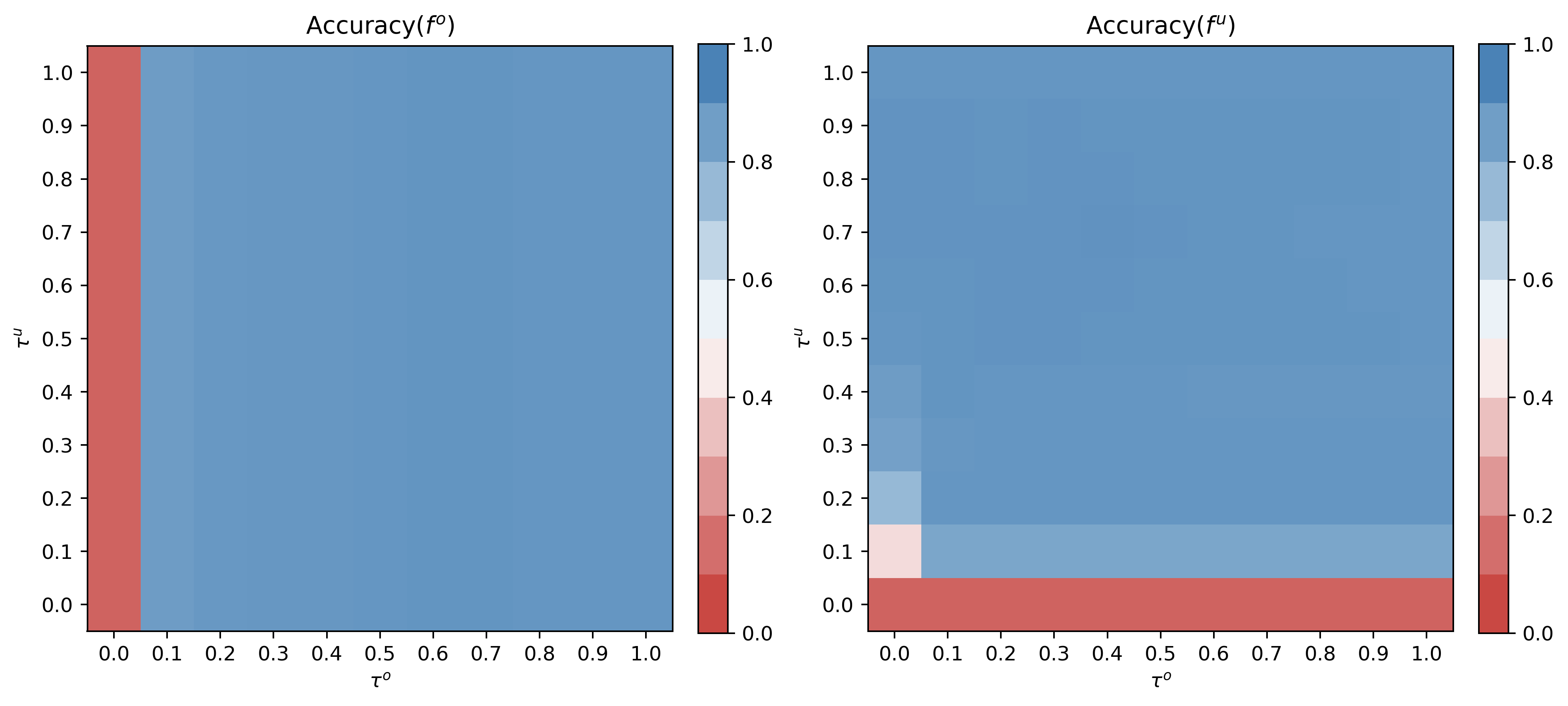}
    \caption
    {Held-out Evaluation Accuracy for $\tau^o$-$\tau^u$ Pairs.
    Original models display good accuracy ($Accuracy(f^o) > 0.7$) when $\tau^o>0.1$.
    Updated models display good accuracy when $\tau^u>0.1$.
    }
    \label{fig:btc_sweep_taus_accuracies}
\end{figure}

\begin{figure}[!h]
    \centering
    \includegraphics[width=\columnwidth]{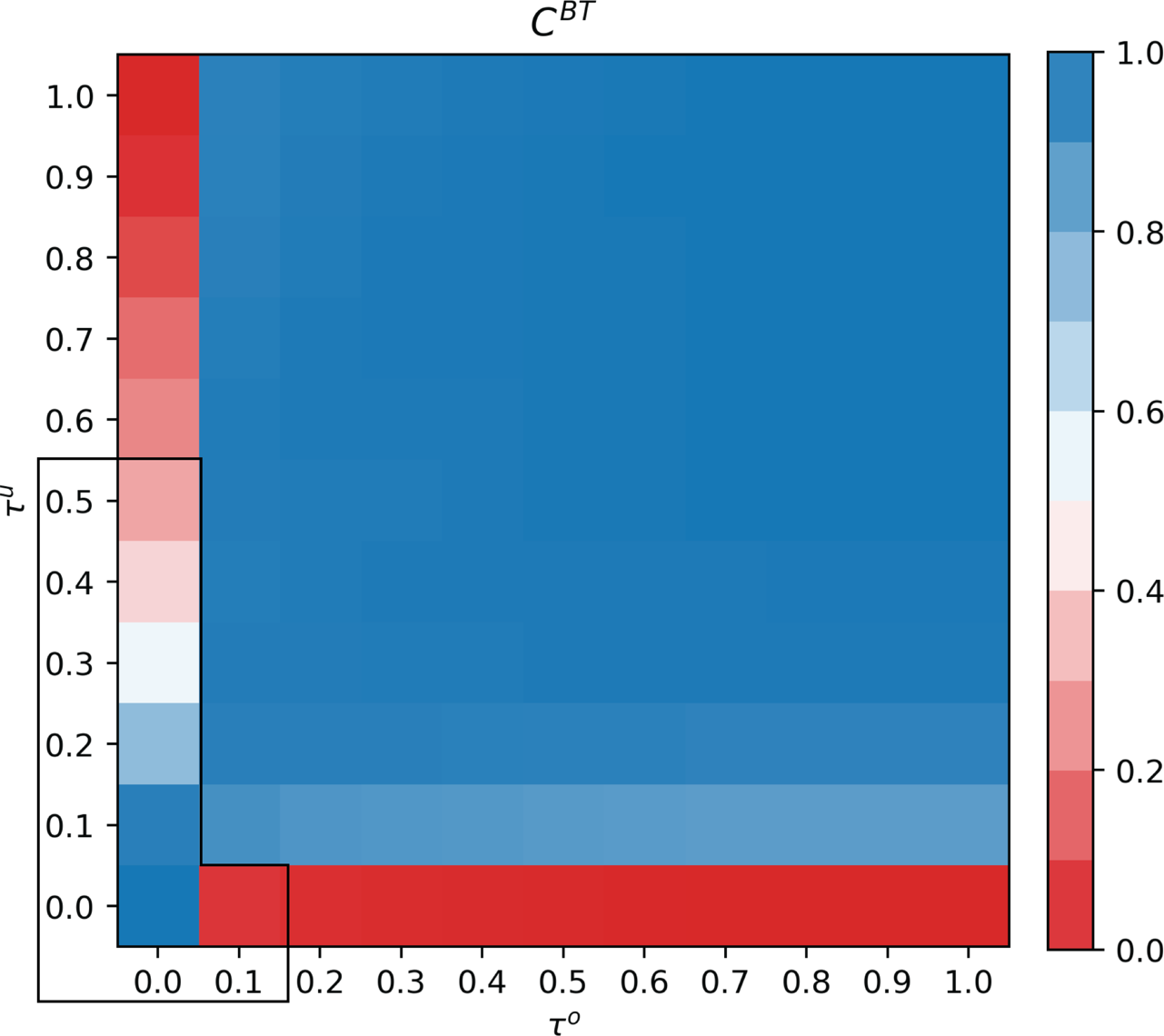}
    \caption
    {Mean Maximum Achievable Held-out Evaluation $\BTC$ for $\tau^o$-$\tau^u$ Pairs.
    The majority of model-pairs yield good maximum achievable $\BTC$ values.
    The boxed area denotes an area of poor performance, depicted in detail in \textbf{Figure \ref{fig:btc_sweep_taus_detail}}.
    }
    \label{fig:btc_sweep_taus_overview}
\end{figure}

For each replication, we swept both $\tau^o$ and $\tau^u$ and selected the updated ``BCE model'' that maximized the validation $\BTC$.
We then computed the $\BTC$ on the held-out evaluation dataset for the selected updated ``BCE model'' given the two threshold values and the original model.
In \textbf{Figure \ref{fig:btc_sweep_taus_accuracies}}, we show the accuracy of each model, and in \textbf{Figure \ref{fig:btc_sweep_taus_overview}}, we show the mean evaluation $\BTC$ for each $\tau^o$-$\tau^u$ pair.

\begin{figure}[!h]
    \centering
    \includegraphics[height=8in]{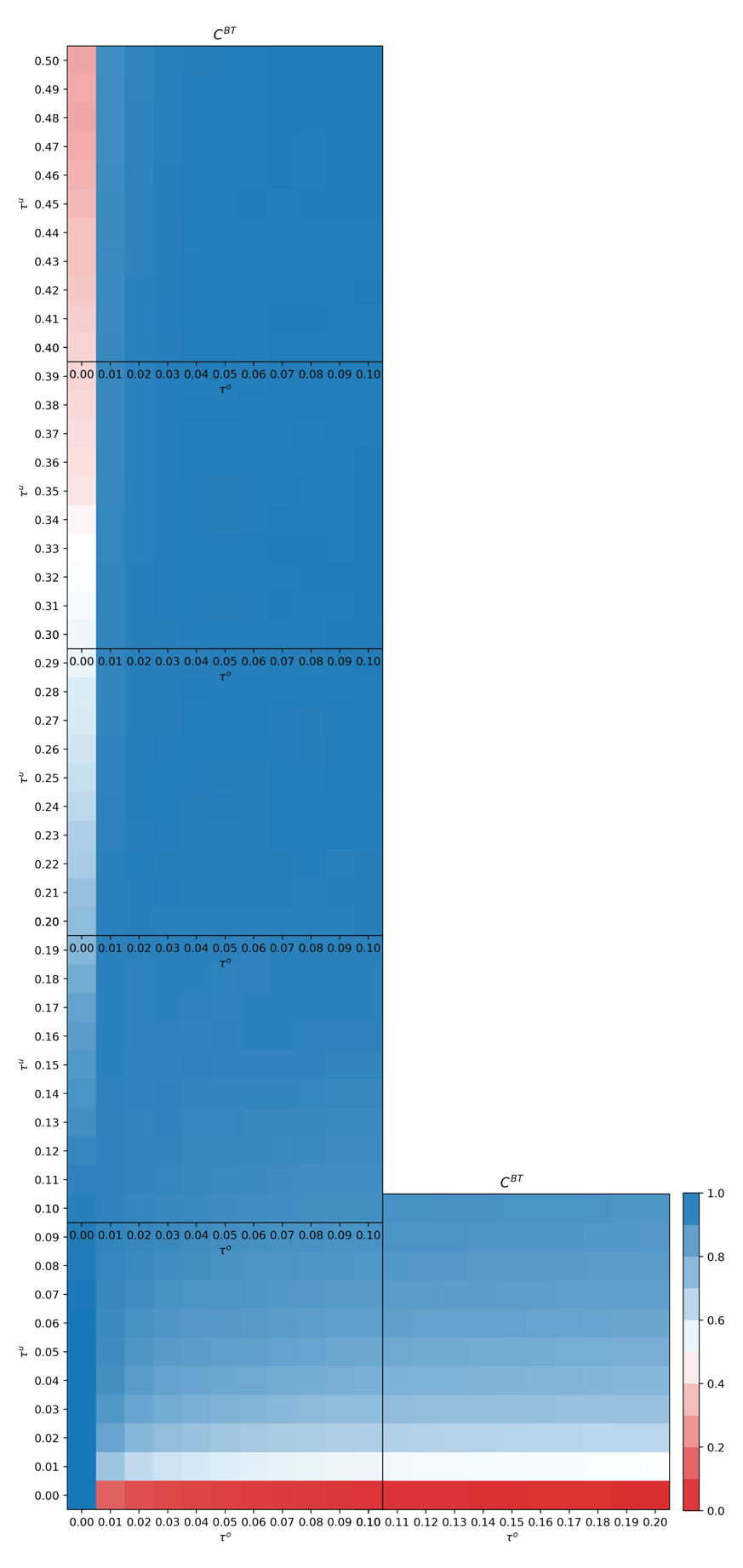}
    \caption
    {Detailed View of Mean Maximum Achievable Held-out Evaluation $\BTC$ for $\tau^o$-$\tau^u$ Model-Pairs.
    }
    \label{fig:btc_sweep_taus_detail}
\end{figure}

We note that many of $\tau^o$-$\tau^u$ pairs corresponding to model-pairs with good accuracy (\ie $\tau^o, \tau^u \geq 0.1$) yield good maximum achievable $\BTC$ values ($\BTC(f^o, f^u) > 0.9$).
An area of interest is where $\tau^o$ is very low (between $0$ and $0.1$) and where $\tau^o$ is low, and $\tau^u$ is very low ($ 0.1 \leq \tau^o \leq 0.2$ and $0 \leq \tau^u \leq 0.1$).
We show fine-grained results for this area in \textbf{Figure \ref{fig:btc_sweep_taus_detail}}.
In this detailed view, we see that poor $\BTC$ values ($\BTC(f^o, f^u) < 0.5$) are achieved under two conditions.
The first is when $\tau^o=0$ and $\tau^u \geq 0.35$.
In this case, the $\BTC$ value decreases as $\tau^u$ increases.
The second is when $\tau^o \geq 0$ and $\tau^u \leq 0.01$.
In this case, lower $\tau^u$ values correspond with lower $\BTC$ values.

These extreme case threshold values cause the models to tip their classification balances against one another, \eg the original model and decision threshold may label everyone as $0$, and the updated model and threshold label everyone as $1$.
We note that the regions of large variation in $\BTC$ only occurs in areas where one of the models has bad accuracy and are an empirical observation of the illustrative example depicted in \textbf{Appendix Section \ref{sec:decision_threshold_dependence_of_btc}}.
Additionally, the values in these figures cannot be directly compared to the $\RBC$.
However, they underscore the dependence of $\BTC$ on decision thresholds.

\newpage

\end{document}